\documentclass[journal]{IEEEtran}
\usepackage{amsmath,amsfonts}
\usepackage{algorithmic}
\usepackage{array}
\usepackage{epsfig}
\usepackage{graphicx}
\usepackage{cite}
\usepackage{xcolor}
\usepackage{multirow}
\usepackage{bbding}
\usepackage{amssymb}
\usepackage{pifont}
\usepackage{caption}
\usepackage{textcomp}
\usepackage{stfloats}
\usepackage[T1]{fontenc}
\usepackage{float}

\usepackage{booktabs} 

\newcommand{\JQ}[1]{\textcolor{black}{#1}}

\definecolor{rblue}{rgb}{0,0.5,1}
\definecolor{awesome}{rgb}{1.0, 0.13, 0.32}
\definecolor{hollywoodcerise}{rgb}{0.96, 0.0, 0.63}
\definecolor{lasallegreen}{rgb}{0.03, 0.47, 0.19}
\definecolor{hanpurple}{rgb}{0.32, 0.09, 0.98}
\definecolor{green(pigment)}{rgb}{0.0, 0.65, 0.31}
\usepackage[pagebackref=false,breaklinks=true,colorlinks,bookmarks=false]{hyperref}
\hypersetup{colorlinks=true,linkcolor={red},citecolor={hanpurple},urlcolor={magenta}}
 
\def\ie{\textit{i.e}.}

\begin{document}

\title{Towards Single-Lens Controllable Depth-of-Field Imaging via Depth-Aware Point Spread Functions}

\author{Xiaolong Qian$^{1,}\IEEEauthorrefmark{1}$, Qi Jiang$^{1,}\IEEEauthorrefmark{1}$, Yao Gao$^{1}$, Shaohua Gao$^{1}$, Zhonghua Yi$^{1}$, Lei Sun$^{1}$,\\Kai Wei$^{1}$, Haifeng Li$^{1}$, Kailun Yang$^{2,3,}$\IEEEauthorrefmark{2}, Kaiwei Wang$^{1,}$\IEEEauthorrefmark{2}, and Jian Bai$^{1}$
\thanks{This work was supported in part by the Zhejiang Provincial Natural Science Foundation of China under Grant No. LZ24F050003, the Henan Province Key R\&D Special Project (231111112700), the National Natural Science Foundation of China (NSFC) under Grant No. 62473139, and Hangzhou SurImage Technology Company Ltd.}%
\thanks{$^{1}$X. Qian, Q. Jiang, Y. Gao, S. Gao, Z. Yi, L. Sun, K. Wei, H. Li, K. Wang, and J. Bai are with the State Key Laboratory of Extreme Photonics and Instrumentation, Zhejiang University, Hangzhou 310027, China (e-mail: wangkaiwei@zju.edu.cn).}%
\thanks{$^{2}$K. Yang is with the School of Robotics, Hunan University, Changsha 410012, China (e-mail: kailun.yang@hnu.edu.cn).}%
\thanks{$^{3}$K. Yang is also with the National Engineering Research Center of Robot Visual Perception and Control Technology, Hunan University, Changsha 410082, China.}%
\thanks{\IEEEauthorrefmark{1}Equal contribution.}
\thanks{\IEEEauthorrefmark{2}Corresponding author: Kaiwei Wang and Kailun Yang.}%
}

\markboth{IEEE Transactions on Computational Imaging, February~2025}%
{Qian \MakeLowercase{\textit{et al.}}: Single-lens Depth-of-Field Imaging}
%

\maketitle

\begin{abstract}
Controllable Depth-of-Field (DoF) imaging commonly produces amazing visual effects based on heavy and expensive high-end lenses. However, confronted with the increasing demand for mobile scenarios, it is desirable to achieve a lightweight solution with Minimalist Optical Systems (MOS). This work centers around two major limitations of MOS, \ie, the severe optical aberrations and uncontrollable DoF, for achieving single-lens controllable DoF imaging via computational methods. A Depth-aware Controllable DoF Imaging (DCDI) framework is proposed equipped with All-in-Focus (AiF) aberration correction and monocular depth estimation, where the recovered image and corresponding depth map are utilized to produce imaging results under diverse DoFs of any high-end lens via patch-wise convolution. To address the depth-varying optical degradation, we introduce a Depth-aware Degradation-adaptive Training (DA$^{2}$T) scheme. At the dataset level, a Depth-aware Aberration MOS (DAMOS) dataset is established based on the simulation of Point Spread Functions (PSFs) under different object distances. Additionally, we design two plug-and-play depth-aware mechanisms to embed depth information into the aberration image recovery for better tackling depth-aware degradation. Furthermore, we propose a storage-efficient Omni-Lens-Field model to represent the 4D PSF library of various lenses. With the predicted depth map, recovered image, and depth-aware PSF map inferred by Omni-Lens-Field, single-lens controllable DoF imaging is achieved. To the best of our knowledge, we are the first to explore the single-lens controllable DoF imaging solution. Comprehensive experimental results demonstrate that the proposed framework enhances the recovery performance, and attains impressive single-lens controllable DoF imaging results, providing a seminal baseline for this field. The source code and the established dataset will be publicly available at \url{https://github.com/XiaolongQian/DCDI}.

\end{abstract}


\section{Introduction}
\IEEEPARstart
{I}{n} photography, an All-in-Focus (AiF) image may not effectively convey the creator's intent.
Professional photographers often use Depth of Field (DoF), which refers to the range of depths in a scene that is imaged sharply in focus, as a composition tool to achieve amazing artistic effects. 
This range is mainly determined by the aperture of the camera lens used for shooting: a narrow aperture produces a wide DoF, and a wide aperture produces a shallow DoF~\cite{freeman1990optics}. 
In portraiture, for example, a shallow DoF and strong background bokeh enable the photographer to highlight a subject. 
However, achieving controllable DoF requires a well-designed high-end lens, which is costly, inconvenient, and often challenging to operate, limiting their widespread use in low-budget scenarios or on low-payload platforms~\cite{davis2011creative,correll2020digital}.
An interesting question arises: Is it feasible to achieve controllable DoF imaging using only a Minimalist Optical System (MOS), such as a single-lens optical system?
Due to the lack of necessary optical optimization variables, MOS suffers from low imaging quality and an inflexible DoF. 
Yet, the scene information can be reconstructed by predicting the AiF aberration-free image and the corresponding depth map of the aberration image, making our goal attainable.
Recently, the great success of deep-learning-based methods have promoted the progress of Monocular Depth Estimation (MDE)~\cite{eigen2014depth,lee2019big,bhat2021adabins,yuan2022neural,piccinelli2023idisc,guizilini2023towards,yin2023metric3d,piccinelli2024unidepth} and Computational Aberration Correction (CAC)~\cite{peng2016diffractive,chen2021optical,hu2021image,jiang2022annular,jiang2023minimalist,jiang2024realworld_cac}. 

\begin{figure*}[!t]
    \centering
    \includegraphics[width=1.0\linewidth]{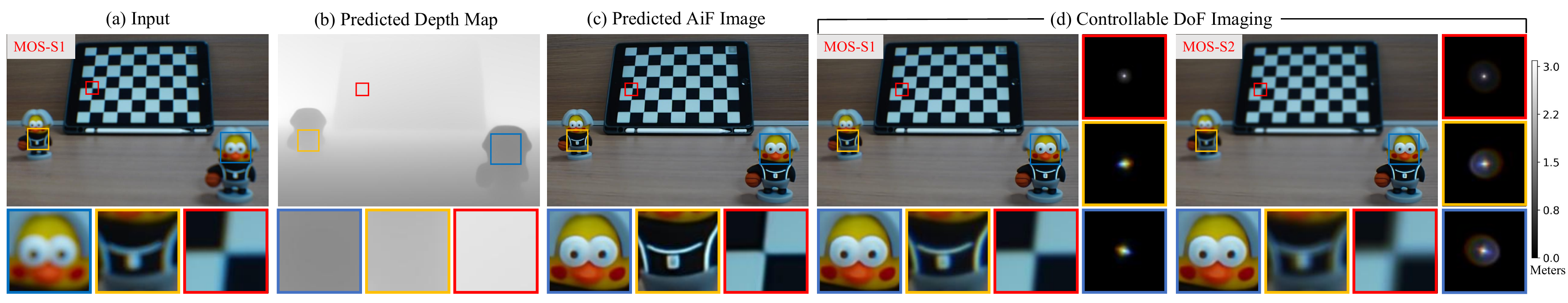}
    \caption{Post-capture controllable Depth-of-Field (DoF) Imaging. (a)~Aberration image captured by the MOS-S1; (b)~Predicted depth map from UniDepth~\cite{piccinelli2024unidepth}; (c)~Predicted All-in-Focus (AiF) image recovered by our Depth-aware Aberration Correction Network (DACN); (d)~Controllable DoF imaging. Our method can produce diverse DoF imaging effects of different lenses. We show our real-world results, keeping the nearest doll in the scene clear while generating bokeh effects of two different lenses (MOS-S1 and MOS-S2). Detail regions are highlighted at different depths and the corresponding PSFs, which are amplified with colored boxes. Please zoom in for the best view.}
    \label{fig:Dof control}
\end{figure*}

For estimating scene depth, researchers have demonstrated that the depth-aware aberration can encode the depth information into the optical degradation, contributing to more accurate monocular depth estimation~\cite{chang2019deep,ikoma2021depth,tan20213d,liu2022investigating}.
As for CAC, the majority of studies focus on processing aberration images from simple, imperfect optical lenses~\cite{peng2016diffractive,hu2021image,chen2021optical} or compact imaging systems with a large Field of View (FoV)~\cite{peng2019learned,jiang2022annular,jiang2023minimalist}. 
Nevertheless, these methods ignore the significant property of optical degradation: aberration behaviors vary across different object distances. 
When the object distance, \ie, the scene depth, changes drastically, these depth-uncorrelated methods are unable to produce aberration-free AiF results, suffering severe artifacts.

To address this gap, we propose a Depth-aware Controllable DoF Imaging (DCDI) framework. 
This framework revolves around three tasks: depth estimation of aberration image, CAC, and PSF characterization.
Fig.~\ref{fig:Dof control} shows that our framework can predict the AiF aberration-free image and the corresponding depth map based on only one aberration image.
By using these two images and the PSF representation module, controllable DoF imaging effects for different lenses are achieved.

For obtaining scene information, a generalized metric MDE model~\cite{piccinelli2024unidepth} is used to generate reliable metric depth maps.
To aid in recovering the aberration-free AiF image from the depth-aware optical degradation, we establish the Depth-aware Aberration MOS (DAMOS) dataset by ray tracing the PSF map of different object distances and introduce the Depth-aware Degradation-adaptive Training (DA$^{2}$T) scheme.
Considering that depth information may enable the network to understand the optical degradation process that varies with depth, we design two plug-and-play depth-aware mechanisms: Residual Depth-Image Cross-Attention Block (RDICAB) and Depth-aware Deformable Convolution Block (D$^{2}$CB).
In RDICAB, the cross-attention mechanism is utilized for the initial fusion of depth features and image features, while in D$^{2}$CB, the image feature is modulated by the depth feature through deformable convolution to adaptively learn the depth-aware degradation.

Based on the recovered AiF aberration-free image and corresponding depth map, controllable DoF imaging results can be produced by using a depth-aware PSF map of any lens via patch-wise convolution.
To address the huge storage space consumption of the original 4D PSF Library (PSFLib), we introduce the storage-efficient Omni-Lens-Field model to accurately characterize the 4D PSFLib for various known lens parameters while occupying low storage consumption.

To assess the effectiveness of our proposed DA$^{2}$T scheme and two depth-aware mechanisms, we conduct extensive experiments to test the generalizability of the trained CAC model on various datasets, including both simulated and real-world datasets. 
The quantitative and qualitative experimental results demonstrate that the model trained by the DA$^{2}$T scheme generalizes better than the non-DA$^{2}$T restoration model in terms of CAC.
Specifically, the DA$^{2}$T model successfully restores scenes at different depths, whereas the non-depth-aware model fails to do so. 
The quantitative and qualitative results also prove that under the same DA$^{2}$T scheme, both RDICAB and D$^{2}$CB enable the model to restore more details and reduce artifacts by fusing depth information.
To further verify the generalizability of our method, we demonstrate results on four state-of-the-art Super-Resolution (SR) architectures~\cite{lim2017enhanced,liang2021swinir,zhou2023srformer,Hsu_2024_CVPR}, on two single-lens systems (MOS-S1 and MOS-S2), showing that our approach is not strictly tied to a specific restoration module and can consistently enhance the recovery performance.

The experimental results show that high-quality AiF aberration correction is achieved, providing a favorable precondition for controllable DoF imaging.
By utilizing the restored image, the estimated depth map, and the PSF map inferred from the Omni-Lens-Field, controllable DoF imaging with multiple lenses is achieved.
Considering the lack of Ground Truth (GT) for the controllable DoF imaging, we only present ``qualitative'' results instead of conducting the GT-required ``quantitative'' evaluation.

To enhance readability, Table~\ref{tab:abbreviations} presents all abbreviations and their full forms used in this paper.

\begin{table}[!t]
    \centering
    \caption{Glossary of Terms and Abbreviations}
    \resizebox{0.8\columnwidth}{!}{%
\begin{tabular}{ll}
    \toprule 
    \textbf{Abbreviation} & \textbf{Full Form} \\ 
    \midrule 
    AiF & All in Focus \\
    CAC & Computational Aberration Correction \\
    CoC & Circle-of-Confusion \\
    DAMOS & Depth-aware Aberration MOS \\
    DA$^{2}$T & Depth-aware Degradation-adaptive Training \\
    DACN & Depth-aware Aberration Correction Network \\
    D$^{2}$CB & Depth-aware Deformable Convolution Block \\
    DCDI & Depth-aware Controllable DoF Imaging \\
    DCL & Deformable Convolution Layer \\
    DoF & Depth-of-Field \\
    FoV & Field of View \\
    K & Key \\
    MDE & Monocular Depth Estimation \\
    MLP & Multi-Layer Perceptron \\
    MOS & Minimalist Optical System \\
    PSF & Point Spread Function \\
    PSFLib & PSF Library \\
    PSA & Permuted Self-Attention \\
    Q & Query \\
    RB & Residual Block \\
    RDG & Residual Deep-feature-extraction Group \\
    RDICAB & Residual Depth-Image Cross-Attention Block \\
    RSTB & Residual Swin Transformer Block \\
    V & Value \\
    \bottomrule 
\end{tabular}
}
    \label{tab:abbreviations}
\end{table}

In summary, we deliver the following contributions:
\begin{itemize}
    \item We propose a Depth-aware Controllable DoF Imaging (DCDI) framework, which achieves controllable DoF imaging for MOS through all-in-focus aberration correction and monocular depth estimation.
    \item A Depth-aware Degradation-adaptive Training (DA$^{2}$T) scheme is introduced, where the models trained with this scheme can achieve better CAC results on both simulated and real-world data in MOS.
    \item We design two depth-aware mechanisms which seize depth information in aiding CAC for better addressing depth-varying optical degradation. 
    \item The storage-efficient Omni-Lens-Field model is designed to accurately represent the 4D PSF library of multiple lenses to achieve the controllable DoF effect of different lenses.
\end{itemize}

\section{Related Work}
\subsection{Imaging Simulation for Minimalist Optical System}
Data-driven image restoration algorithms require large amounts of degraded-sharp image pairs for training~\cite{Wang2021SRSurvey}. 
The conventional method of manual data acquisition involves utilizing a MOS alongside a high-end lens to capture the same scene simultaneously, thereby generating a pair of aberrated-sharp images. 
This technique is both time-intensive and demanding in terms of effort, further complicated by intricate challenges associated with aligning images accurately. 
These impediments substantially restrict the progression within the field of CAC. 
Hence, image simulation is often a better choice for simulating synthetic datasets in CAC~\cite{chen2021optical,luo2024correcting,gong2024physics}. 
In addition, the degradation caused by the optical system is typically modeled as the convolution of a sharp image with the PSFs. 
Thus, accurately estimating the PSF of the optical system is imperative, whether the goal is to simulate images that reflect true aberration degradation or to improve the efficacy of aberration correction techniques.
In the visual sensing field, the commonly used idealized optical model~\cite{gur2019single,nazir2023depth} calculates the PSF based on the paraxial principle, generating a truncated Gaussian function called the ``Circle-of-Confusion'' (CoC). The diameter of CoC can be described by 
\begin{equation}
    CoC = \frac{F}{N}\frac{|D-f_d|}{D}\frac{F}{f_d-F},
\end{equation}
where $F$ is the focal length and $N$ is the F-number of the camera lens. 
$D$ is the distance from the object to the lens, also known as the object depth. 
$f_d$ is the focal distance.

However, due to the limitations of the paraxial approximation, the aforementioned ideal model falls short of accurately representing real-world optical systems, which exhibit off-axis aberrations. 
A widely employed technique, PSF calibration~\cite{li2019depth,chen2020data}, entails gauging a lens's reaction to a designated point light source, thereby facilitating the estimation of the PSF at unassessed locations. 
Recently, the technique has seen significant advancement through applying low-rank decomposition~\cite{yanny2020miniscope3d,yanny2022deep}, which models the spatially varying PSFs as a weighted combination of shift-invariant kernels. 
Ray tracing~\cite{sun2021end,chen2021optical,wang2022differentiable,cote2023differentiable}, as a direct method for PSF calculation, can obtain accurate spatially-varying PSF which has been widely used in commercial software such as ZEMAX and Code V.

Unfortunately, the aforementioned PSF estimation methods typically only consider a single object distance, ignoring the effect of scene depth on the PSF. 
Luo~\textit{et al.}~\cite{luo2024correcting} rethink the imaging process of realistic 3D scenes by designing an effective simulation system to compute muli-wavelength, spatially-variant, depth-aware four-dimensional PSFs (4D-PSFs), which can, to some extent, better approximate the actual physical degradation model and bridge the gap between simulation and real-world imaging.

For the first time, we consider 4D-PSFs in the simulation process of MOS and propose the Depth-aware Degradation-adaptive Training (DA$^{2}$T) scheme. 
Based on the proposed DA$^{2}$T scheme, CAC models can achieve high-quality AiF aberration correction on the established dataset.

\subsection{Computational Aberration Correction}
Compactly designed, simple optical systems are inherently flawed, leading to sub-optimal images with aberrations. 
The Computational Aberration Correction (CAC) methods are widely used as a post-image processing module to improve the image quality of these systems, which is a classic computational imaging framework~\cite{barbastathis2019use}.

Most classic restoration methods~\cite{joshi2008psf,kee2011modeling,schuler2011non,schuler2012blind,heide2013high} attribute image degradation to a convolution process. 
In this way, they restore degraded data by estimating the degraded kernel. 
However, the effectiveness of these model-based methods relies on the accuracy of the estimated degraded kernel and suffers from poor generalization performance. In the last decade, deep-learning-based methods have been widely employed in image restoration~\cite{sun2015learning,chen2021hinet,chen2022simple}, and the CAC area has also made great progress, achieving remarkable restoration results. 
These methods typically use an encoder-decoder network structure to fit the degradation process of the optical system through reverse gradient optimization~\cite{peng2019learned,hu2021image}. Diverging from purely data-driven algorithms, some studies~\cite{chen2021extreme,jiang2022annular,jiang2023minimalist,chen2023mobile,luo2024correcting} investigate the integration of spatial-variant optical prior knowledge, such as the FoV and PSF cues, into the network to enhance image restoration. 
Additionally, some researches~\cite{ruan2021aifnet,ruan2022learning} have proposed defocus deblurring methods that account for depth-varying PSFs; however, they often overlook the aberrations introduced by imperfections in the optical system design.

To address the degradation caused by real-world lenses, Luo et al.~\cite{luo2024correcting} estimated the depth map of the aberration image and selected the corresponding PSF from PSFLib to restore the aberration image. 
However, their method did not directly use depth information for aberration correction or other tasks.
In contrast, our framework leverages the estimated depth from the aberration image to enhance CAC performance, addressing depth-aware degradation and enabling effective restoration of scenes with varying depths. 
Moreover, we explore the use of estimated depth maps and restored images for other applications, such as controllable DoF imaging.

\subsection{Controllable Depth-of-Field Imaging}
Earlier work on DoF manipulation is achieved through hardware~\cite{Dowski:95,george2003extended,hong2007flexible,kuthirummal2010flexible}.
Hong~\textit{et al.}~\cite{hong2007flexible} propose a method using a spatial light modulator at the pupil plane to control the DoF in optical microscopes flexibly. 
Kuthirumma~\textit{et al.}~\cite{kuthirummal2010flexible} design a camera system with flexible DoF by controlling the motion of the detector during the image integration process. 
The rapid advancements in deep learning have led to the emergence of a growing number of learning-based approaches for controlling DoF. 
With the availability of an input or an estimated depth map, a shallow DoF image can be synthesized through either a neural network-based approach~\cite{wang2018deeplens,wadhwa2018synthetic} or a hybrid method combining classical and neural rendering techniques~\cite{peng2022bokehme}. 
In addition, one method to achieve controllable DoF imaging is to capture a \JQ{focal stack}, and then composite the relevant frames to emulate the target DoF, but this approach suffers from the lengthy capture time, which restricts its use to static scenes~\cite{jacobs2012focal}. 
Alzayer~\textit{et al.}~\cite{alzayer2023dc2} leverage the dual-camera system on the smartphone to achieve various DoF manipulations through defocus control.
However, the optical systems considered in the aforementioned methods are idealized, without accounting for the potential image degradation caused by optical aberrations. 
Therefore, in our work, the proposed DCDI framework enables controllable DoF imaging for the MOS suffering from aberration-induced degradation and uncontrolled DoF. 
It is essential to emphasize that the proposed method can synthesize the imaging effects of any lens with known parameters and control its DoF.

\section{Methodology}
\begin{figure*}[!t]
    \centering
    \includegraphics[width=1.0\linewidth]{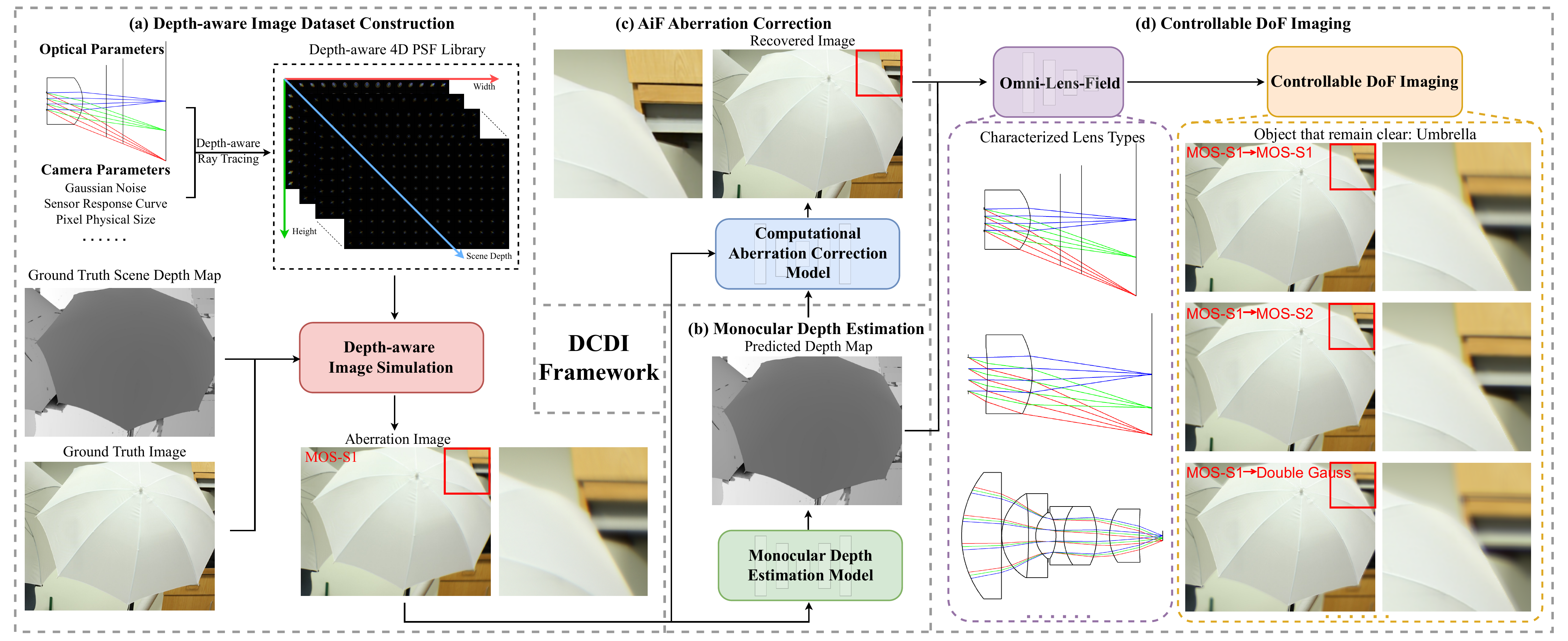}
    \caption{Overview of the proposed Depth-aware Controllable DoF Imaging (DCDI) framework. (a)~We simulate the Depth-aware Aberration MOS (DAMOS) dataset. The Depth-aware 4D PSFLib is constructed by performing multiple ray tracing simulations, varying the position of the object plane from near to far. (b)~The MDE model predicts scene depth map. (c)~The latent AiF image is recovered by jointly combining synthetic data, predicted depth map, and depth-aware network architecture. (d)~Controllable DoF imaging of multiple lenses is achieved through predicted depth maps, restored AiF images, and depth-aware PSF maps predicted by Omni-Lens-Field.}    
    \label{fig:framework}
\end{figure*}

The proposed DCDI framework is illustrated in Fig.~\ref{fig:framework}. 
Our framework is comprised of four components: the simulation of a depth-aware aberration dataset, an MDE module, a CAC module, and a PSF representation module.
In Sec.~\ref{section:dataset}, we introduce a method for simulating a depth-aware aberration dataset, which serves as a foundation for enabling the network to adaptively learn depth-varying degradation characteristics during Depth-aware Degradation-adaptive Training (DA$^{2}$T) scheme.
In Sec.~\ref{section:network}, Residual Depth-Image Cross-Attention Block (RDICAB) and Depth-aware Deformable Convolution Block (D$^{2}$CB) are proposed as two depth-aware mechanisms to enhance the recovery performance of the Depth-Aware Correction Network (DACN). 
In Sec.~\ref{section:dof_imging}, we design the Omni-Lens-Field to represent the 4D PSFLib of various lenses. 
Based on the recovered AiF aberration-free image, the depth map estimated by UniDepth~\cite{piccinelli2024unidepth} and the depth-aware PSF map inferred by Omni-Lens-Field, single-lens controllable DoF imaging is achieved.  

\subsection{Depth-aware Image Dataset Construction}
\label{section:dataset}

In this part, we introduce how to build a depth-aware aberration image dataset, including the construction of the PSFLib and image simulation. 
The construction of the depth-aware image dataset, as a key component, is integrated into the training process with the aim of embedding depth-varying optical characteristics into network training. A detailed introduction to DA$^{2}$T is provided in the appendix.

\textbf{PSFLib establishment.} 
The energy dispersion of the PSF can typically represent the aberration degradation caused by the incompleteness of the MOS. 
Therefore, we employ a ray tracing model~\cite{sun2021end,wang2022differentiable,cote2023differentiable} which has been extensively experimentally proven to accurately obtain spatially-varying PSFs. 
The precise implementation of ray tracing involves an iterative process comprising two key steps: 
1) updating the ray coordinates from one optical interface to the next interface using the Newton iteration method, and 
2) updating the ray direction cosines in accordance with Snell's law of refraction. 
The ray tracing applies a ray-aiming correction step~\cite{cote2023differentiable} to accurately simulate optical systems, particularly those affected by pupil aberrations. 
For systems with large geometric aberrations, the diffraction effect can be ignored~\cite{luo2024correcting,gao2024global}, so the PSF can be considered as the superposition of ray energy traced from the entrance pupil to the image plane, with the energy distribution described by a Gaussian function~\cite{li2021end}:
\begin{equation}
    e_{m,n}^{i,j} =\frac{1}{\sqrt{2 \pi} \sigma} \exp \left(-\frac{(r_{m,n}^{i,j})^{2}}{2 \sigma^{2}}\right).
\end{equation}
Here, $r_{m,n}^{i,j}$ represents the distance between the ray with pixel index $(m,n)$ and the central ray of the image plane, where $(i,j)$ is the index of the ray in the pupil.

We trace $128{\times}128$ rectangular distributed rays in the pupil in practice. 
$\sigma{=}\sqrt{\Delta h^{2}{+}\Delta w^{2}}{/}3 $, and ${\Delta}h{\times}{\Delta}w$ is the physical size of each pixel.
In the standard deviation formula, the number $3$ is chosen as the magic number based on the $3{\sigma}$ rule, which ensures that $99.7\%$ of the light energy is concentrated within ${\pm}\sigma$, allowing us to limit the calculation area to one pixel rather than integrating over an infinite range. This method greatly reduces the computational complexity while maintaining the high accuracy of the results.
After obtaining the PSFs for the sampled field of view and wavelength, we can then synthesize the three-channel RGB PSFs by simulating the spectral response curve of the CMOS sensor~\cite{gao2024global}. 
The three-channel RGB PSF for the same imaging depth can be obtained by the following formula:
\begin{equation}
    \label{PSF}
    PSF(c,\theta,d )=\sum_{\lambda_{s}}^{}R(\lambda_{s})\cdot PSF (c,\lambda_{s},\theta ,d), 
\end{equation}
where $c$ represents one of R, G, and B channels, $\theta $ represents the sampled field of view, which is obtained by dividing the maximum image height of the optical system into $N_{fov}$ parts, $d$ represents the imaging depth, and $R(\lambda_{s})$ represents normalized wavelength response coefficient. 
It is worth noting that while the focal length does affect the PSF as it represents the optical system's ability to converge light, it is a fixed parameter for a given fixed-focus optical system. In Eq.~\ref{PSF}, we aim to emphasize the impact of variations in Field of View (FoV), channel, and object distance on the PSF under the condition of a fixed focal length. Therefore, it is not explicitly expressed in the equation.
To trade off the accuracy and speed of the simulation, the $N_{fov}$ is set to $64$, and the RGB PSF size is set to $3{\times}41{\times}41$.
The common optical systems used in industry, including our MOS, are axisymmetric, resulting in PSFs that don't vary with azimuth.
Therefore, the PSF map can be efficiently filled by interpolating, rotating, and resizing $PSF(c,\theta,d )$. 
The RGB PSF map at a scene depth can be described as:

\begin{align}
PSF \ map(c,h,w,d) = 
&P_{resize} \circ P_{rot}
\notag
\\
&\circ P_{inter}(\theta) \circ PSF(c,\theta,d ),
\end{align}
where $\circ$ represents the composition operator, and $P_{resize}$, $P_{rot}$, and $P_{inter}$ represent the process of resizing, rotating, and interpolating, respectively. $h$ and $w$ represent the patch's position on the image plane. 
Following~\cite{luo2024correcting}, to balance simulation accuracy and storage space, the image is divided into $N_h{\times}N_w$ patches, each with $m{\times}m$ pixels, where the PSF is assumed spatially uniform within each patch, as the PSF exhibits little variation in a small spatial range.

\begin{figure*}[!t]
    \centering
    \includegraphics[width=0.9\linewidth]{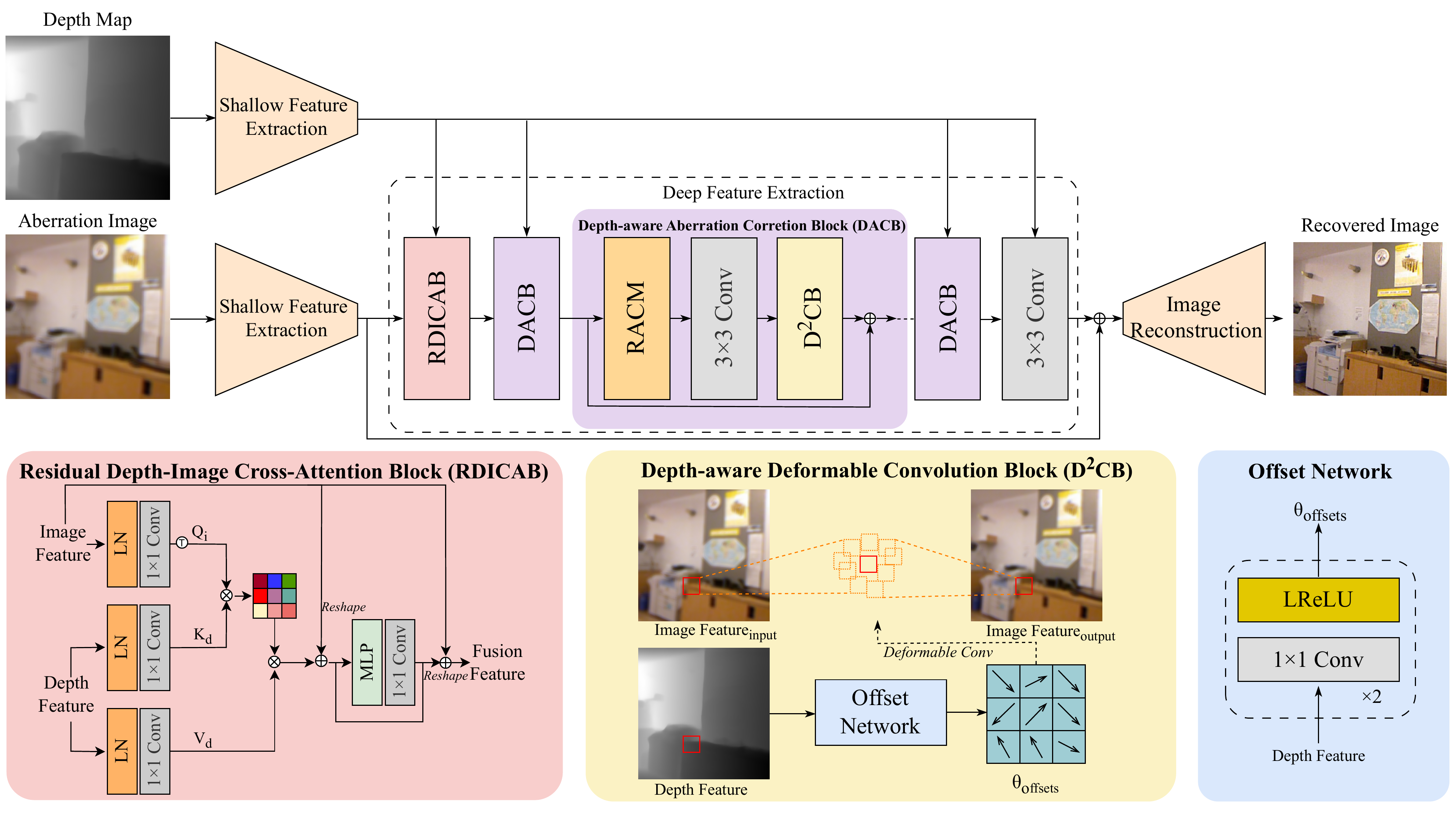}
    \caption{Overview of the proposed Depth-aware Aberration Correction Network (DACN). DACN is established on a classical super-resolution paradigm, which includes shallow feature extraction, deep feature extraction, and reconstruction. RDICAB and D$^{2}$CB are two proposed plug-and-play depth-aware mechanisms that can effectively learn depth-aware degradation with the guidance of depth features. Here, RACM denotes the Replaceable Aberration Correction Module, which implies that any suitable restoration module can be utilized.} 
    \label{fig:Depth-aware Aberration Correction Network}
\end{figure*}

To subsequently construct a depth-aware aberration image dataset, the RGB PSFs obtained through the ray tracing process need to consider the imaging scene at multiple depth planes. 
For this purpose, we perform ray tracing by varying the distance to the object plane and establish a comprehensive PSFLib for the lens, which can be expressed by the following formula:
\begin{align}
PSFLib(c,h,w,d) 
= \left \{ PSF\ map(c,h_{i},w_{i},d_{i}) \mid 
\notag \right.
\\ 
\phantom{=\;\;}
\left. 
0 \le h_{i}  \le N_{h},0 \le w_{i}  \le N_{w} ,d_{min} \le d_{i}  \le d_{max} \right \} ,
\end{align}  
where $d_{i} {\in} \left [ d_{min},d_{max} \right ] $ denotes the scene depth range. 
We set $d_{min}{=}0.7m$ and $d_{max}{=}10.0m$, which is enough to cover the depth range of the training data.

\par \textbf{Depth-aware Image Simulation.} To mimic the image degradation process, patch-wise convolution is applied to simulate the aberration image to deal with the spatially varying PSFs. 
Different from~\cite{cote2023differentiable,gao2024global}, which neglect the imaging scene depth in the simulation process, we take the scene depth into account in our pipeline. 
This is because the imaging scene depth directly affects the degradation of the optical system, and the consideration of this factor helps our simulation process better reflect the actual degradation process.
The depth-aware aberration image is approximately generated by
\begin{equation}
I_{L} =\sum_{h_{i}}^{} \sum_{w_{i}}^{} \psi (PSFLib,D_{avg}) \ast I_{H} + N ,    
\end{equation}
where $\ast$ represents the convolution operation, $I_{H}$ and $I_{L}$ denote ground truth and aberration image respectively, $\psi$ denotes the searching function that queries the PSFLib and outputs the corresponding PSF of every patch with different scene depths. 
\JQ{Here, we also consider the distortion of the optical lens, and perform distortion processing on both the aberration image and the clear image to ensure pixel alignment.
Since distortion does not affect the image quality, distortion correction is not within the scope of this paper.}
The depth value of each patch is the average depth value for all pixels within that patch. 
The average depth map of the entire scene can be expressed as follows
\begin{equation}
    D_{avg} = AvgPool2d(D_{s},m),
\end{equation}
where $D_{avg}$ and $D_{s}$ denote the average scene depth map and primary scene depth map respectively, and $m$ denotes the patch size. 

\subsection{Depth-aware Aberration Correction Network}
\label{section:network}
As demonstrated in~\cite{jiang2023minimalist}, the SR paradigm achieves better results in aberration image recovery.
It should be emphasized that the image resolution is not improved; instead, the pixel unshuffle operation is applied to the input images and depth maps to ensure consistency between input and output resolutions.
For this reason, the proposed Depth-aware Aberration Correction (DACN) is an end-to-end CAC model based on the SR paradigm, enabling deep multimodal fusion of image and depth information.
The architecture of the proposed model is shown in Fig.~\ref{fig:Depth-aware Aberration Correction Network}, which includes our proposed two plug-and-play depth-aware mechanisms. 
Residual Depth-Image Cross-Attention Block (RDICAB) follows the shallow feature extraction layer. 
It is employed for the initial stage of cross-modal fusion, aiming to integrate image features and deep features.
Depth-aware Deformable Convolution Block (D$^2$CB) is employed within each depth-aware aberration correction block. 
The image features are modulated by the depth features through a Deformable Convolution Layer (DCL)~\cite{zhu2019deformable}, which effectively facilitates the learning of the depth-aware aberration degradation process.

\par \textbf{Residual Depth-Image Cross-Attention Block (RDICAB).} 
As illustrated in~\cite{yang2022self}, depth information has been proven to be effective in image dehazing because haze varies with depth. Similarly, based on our observation that image degradation caused by optical aberrations also varies with depth, we propose a Residual Depth-Image Cross-Attention Block (RDICAB) to fully exploit the complementary characteristics of multimodal information. 
Unlike the conventional self-attention mechanism~\cite{vaswani2017attention}, in which Query ($Q$), Key ($K$), and Value ($V$) are all calculated by the same branch of the network, our RDICAB makes use of the cross-attention mechanism between image and depth to perform the initial fusion. 
Specifically, the queries are obtained by the image branch, with keys and values calculated by the depth branch, as shown in the lower left of Fig.~\ref{fig:Depth-aware Aberration Correction Network}. 
Note that the input image feature $I_{i}$ and input depth feature $D_{i}$ are first fed into a normalization layer and a $1{\times}1$ convolution layer, followed by the cross-attention operation, formulated as:
\begin{equation}
    Attention(Q_{i},K_{d},V_{d})=V_{d}\ Softmax(\frac{Q_{i}^{T} K_{d}}{\sqrt{d_{k}} } ) ,
    \label{eq:corss att}
\end{equation}
where $(\cdot )^{T} $ denotes the transpose operator. 
The output of the cross-attention operation is added to $I_{i}$, and then the result of the addition is processed by a Multi-Layer Perceptron (MLP) and a $1{\times}1$ convolution layer. Finally, a long skip residual connection is used to add $I_{i}$ to the intermediate features to obtain final output fusion features $F_{o}$.
Overall, RDICAB, utilizing the cross-attention mechanism, effectively handles the initial fusion of multimodal features and provides high-quality modulated image features for subsequent modules.

\textbf{Depth-aware Deformable Convolution Block (D$^2$CB).} 
Due to the irregular shape and size of spatial-variant PSFs, traditional convolution operations, which rely on fixed positional features around the center, are unable to effectively handle these degradations.
In contrast, deformable convolution can adapt the shape of the convolution kernel by introducing offsets. 
This flexibility allows depth information, which encompasses the characteristics of spatial variation, to guide the convolution process, enhancing the model's ability to handle depth-aware optical degradation.
Different from standard deformable convolutional networks~\cite{dai2017deformable}, which calculates offsets based on image feature maps of the same modality, we only calculate offsets through depth features to modulate image features. 
As shown in the lower right of Fig.~\ref{fig:Depth-aware Aberration Correction Network}, D$^2$CB consists of a DCL~\cite{zhu2019deformable} and an offset network. 
The modulated output image feature $I_{o}$ is generated by a DCL that takes the input image feature $I_{i}$ and an offset produced by feeding the depth feature $D_{i}$ into an offset network.
The modulated output $I_{o}$ is formulated as:
\begin{equation}
    I_{o}=F_{DCL}(I_{i},\theta _{offset}),\theta _{offset} = F_{offset}(D_{i}),
\end{equation}
where $F_{DCL}$ represents the deformable convolution layer and $F_{offset}$ represents the offset network.

\subsection{Controllable Depth-of-Field Imaging}
\label{section:dof_imging}
\begin{figure}[!t]
    \centering
    \includegraphics[width=1.0\linewidth]{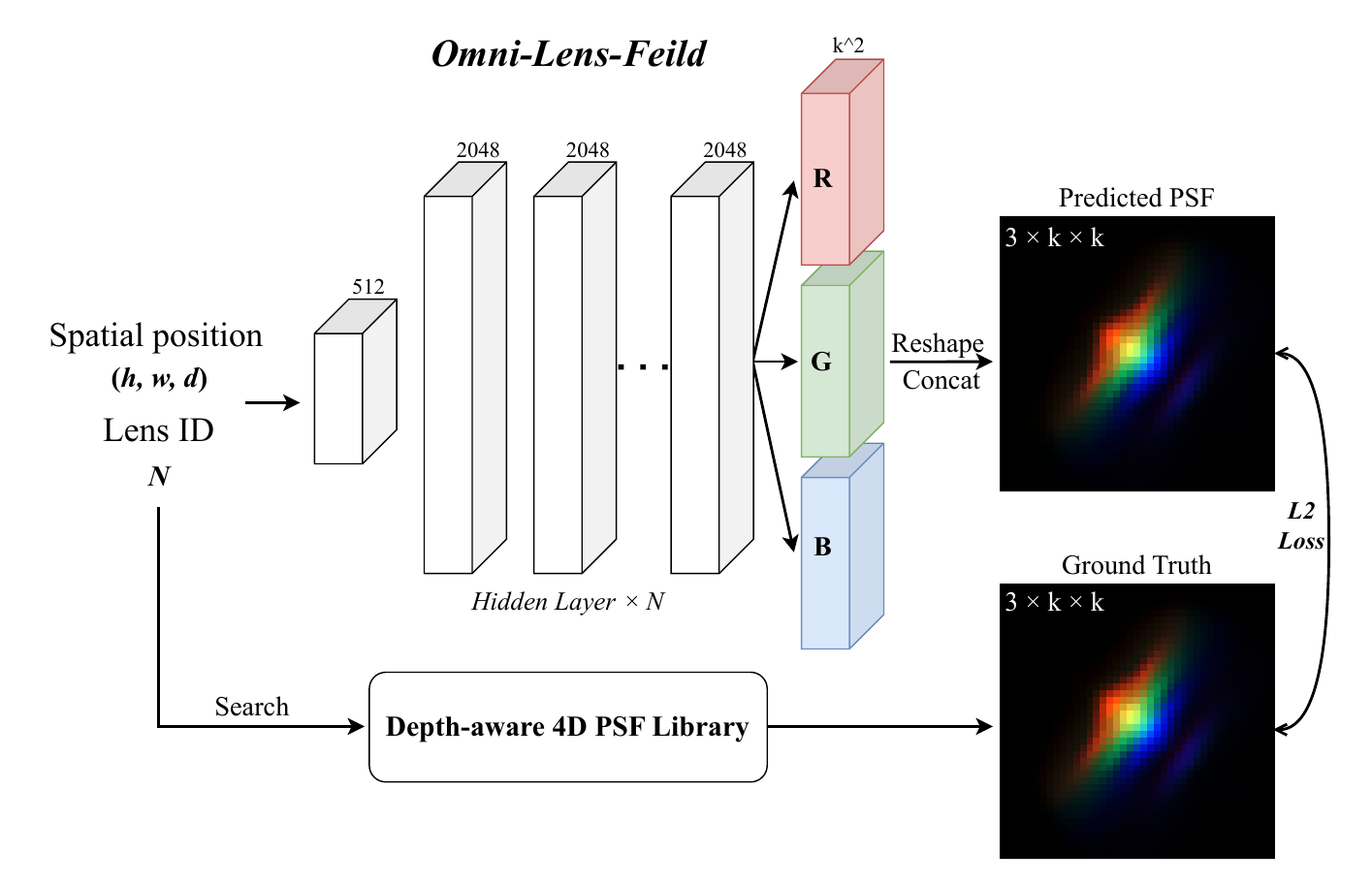}
    \caption{Illustration of the proposed Omni-Lens-Field model training procedure. The network is trained to represent the 4D PSFLib of various lenses. The network takes the normalized coordinates of the image patch on the object plane $(c,h,w)$ and the lens ID $N$ as input and outputs a 3D tensor as the predicted value.} 
    \label{fig:Omni-lens-field}
\end{figure}

To achieve controllable DoF imaging with different lenses, it is necessary to perform ray tracing for each lens individually. 
However, ray tracing is computationally expensive, especially when frequently adjusting the object distance. 
Furthermore, the calculated PSFLib requires large storage space.
To address these challenges, we train the Omni-Lens-Field model to represent the 4D PSFLib for a variety of lenses. 
Different from \cite{tseng2021differentiable,yang2023aberration}, where a network is used to characterize a specific lens, we employ the Omni-Lens-Field model to represent the 4D PSFLib of multiple lenses. 
As illustrated in Fig.~\ref{fig:Omni-lens-field}, we train the network $H_{OLF}$ to fit the PSFs of multiple imaging lenses, by minimizing the discrepancy between the estimated PSF and the ray-traced PSF. 
Following~\cite{yang2023aberration}, L2 Loss is applied during the fitting process, formulated as: 
\begin{equation}
    L(\theta )=\parallel H_{OLF}(h,w,d;N)-PSFLib(h,w,d;N) \parallel _{2}^{2} ,
\end{equation}
where $\theta$ denotes the parameter set of our network, $N$ denotes the lens ID. $(h,w,d)$ represent the normalized coordinates of the image patch on the object plane, where all values are scaled to the range of $\left [ 0,1 \right ]$.

To fit the mapping of the four-parameter input into an RGB PSF, the Omni-Lens-Field adopts an MLP network. 
It consists of one input layer with $512$ channels, $N$ hidden layers with $2048$ channels, and three independent output layers with $k^{2}$ channels. The three independent output layers each predict the PSF values for the corresponding RGB channels. 
The activation function of the input and hidden layers is ReLU, and the Sigmoid is used as the activation function for the three independent output layers. 
The RGB PSF with shape $3{\times}k{\times}k $ can be obtained by reshaping and concatenating the three independent output layers, each with $k^{2}$ channels.

The depth-aware PSF map of any lens can be predicted by Omni-Lens-Field according to the corresponding depth map and can be expressed as follows:
\begin{equation}
    PSF\ Map_{D} = \sum_{}^{} H_{OLF}(h,w,d;N).
    \label{eq:PSF MAP}
\end{equation}

Based on the recovered AiF image $I_{R}$, the predicted depth map $D_{P}$ and the depth-aware PSF map estimated by Omni-Lens-Field, controllable DoF imaging can be achieved. Specifically, through the estimated depth map, we select a depth threshold to keep the objects within the threshold range in the recovered image clear and apply the bokeh effect of different lenses to the scenes outside the threshold range, to achieve controllable DoF imaging of different lenses. The bokeh effect is obtained by convolving the AiF image with the depth-aware PSF map inferred by Omni-Lens-Field. The controllable DoF image $I_{A}$ is formulated as:
\begin{equation}
    I_{A} = \sum_{h_{i}}^{} \sum_{w_{i}}^{} I_{R} \ast \mathit{Mask} (PSF \ Map_{D},\Phi(D_{P})),
    \label{eq:arbitrary dof imaging}
\end{equation}
where $\ast$ represents the convolution operation, $\Phi$ denotes the function that determines the depth threshold based on the subject to keep distinctly visible, the function $\textit{Mask}$ represents the operation that sets the predicted PSF to the identity matrix for the pixels whose depth values fall within the selected threshold range.

\section{Experiments}
\subsection{Datasets}
\begin{figure}[!t]
  \centering
  \includegraphics[width=1.0\linewidth]{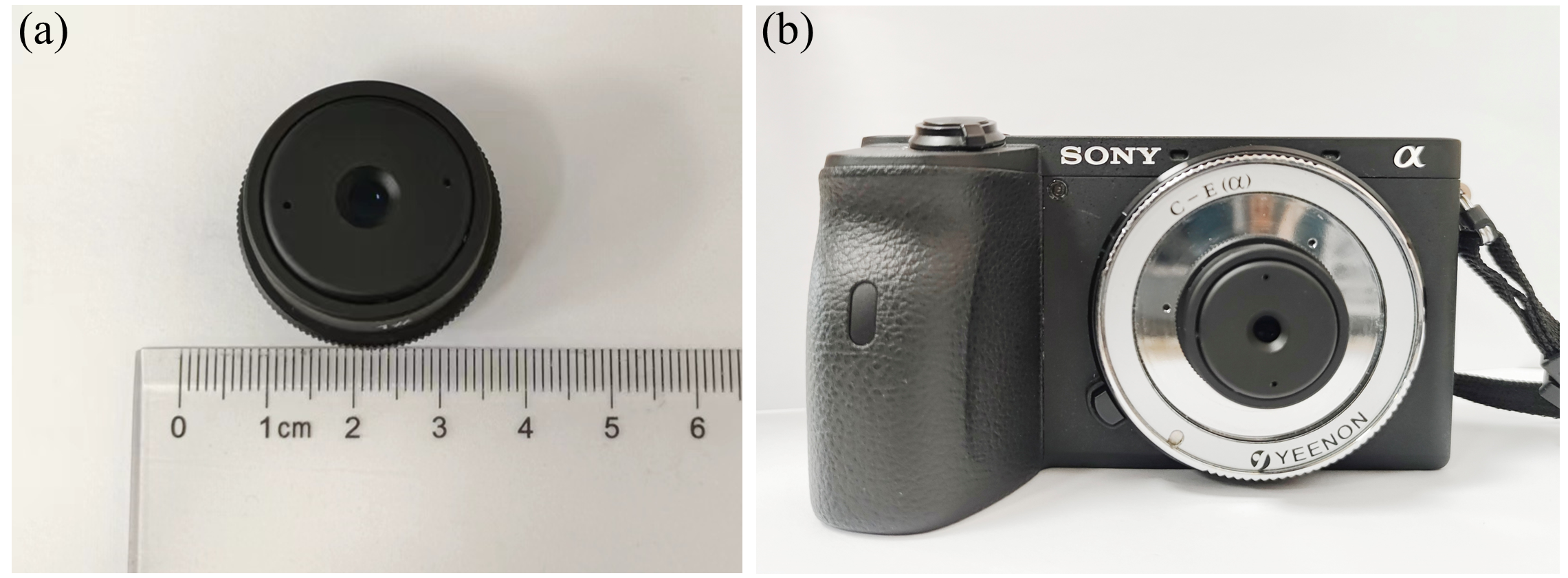}
  \caption{we manufacture (a) MOS-S1 and mount it on a (b) Sony ${\alpha}$6600 camera to capture the real-world dataset.}
  \label{fig:real capture camera}
\end{figure}
\textbf{Synthetic datasets.} 
We establish a Depth-aware Aberration MOS (DAMOS) dataset to train our CAC model, supporting AiF aberration correction.  
The absence of datasets including depth-aware aberration images, corresponding depth maps, and AiF images, prompts us to focus on datasets for CAC consisting of AiF images and depth ground truth and to generate depth-aware aberration images by blurring the AiF images. 
Therefore, we simulate the 4D PSFs on two MOS with distinct aberration distribution (MOS-S1 and MOS-S2) and then use it to perform patch-wise convolution on each color channel of the high-quality AiF images and reunite them to obtain the corresponding synthetic aberration image.
We use the NYUv2 dataset~\cite{silberman2012indoor}, which comprises $1,449$ images of the resolution of $480{\times}640$ as the high-quality AiF images to generate training and test data pairs.
The Middlebury2014 dataset~\cite{scharstein2014high} is also used as the testing dataset, which includes $23$ high-resolution real-world RGB-D images.
The sensor resolution determines the resolution ratio of training and testing images, and the aberration property of the lens can be altered by any resizing operation.
However, the extremely high-resolution images of $4000{\times}6000$ pixels produced by the Sony ${\alpha}$6600 camera sensor cannot be handled during network training processes.
Due to the physical properties of the PSF, downsampling is feasible~\cite{yang2023aberration,luo2024correcting}.
\JQ{Therefore, we downsample the sensor to a $1280{\times}1920$ pixel size using area interpolation, which preserves more image details and reduces the generation of artifacts, considering the resolution of the used datasets.}
For the NYUv2 dataset, we treat the images as part of a $1280{\times}1920$ resolution image to enable the model trained on this dataset to restore the real-world data without any finetuning. 
The resolution of the images in Middlebury2014 is reshaped and cropped to $1280{\times}1920$.
Following the setup in~\cite{jiang2022annular}, we randomly perturb the parameters of the MOS within a range of $10\%$ to generate image pairs with different aberration distributions, which can alleviate the domain gap between synthetic and real-world images.

\textbf{Real-world datasets.} 
As shown in Fig.~\ref{fig:real capture camera}, we manufacture MOS-S1 and mount it on a Sony ${\alpha}$6600 camera to capture real-world scenes containing objects at different depths. 
To assess the generalization capabilities of our proposed method, we evaluate the model trained solely on the NYUv2 dataset~\cite{silberman2012indoor} for performing aberration correction on real-world scenes, without any fine-tuning or additional training.

\begin{figure}[!t]
  \centering
  \includegraphics[width=1.0\linewidth]{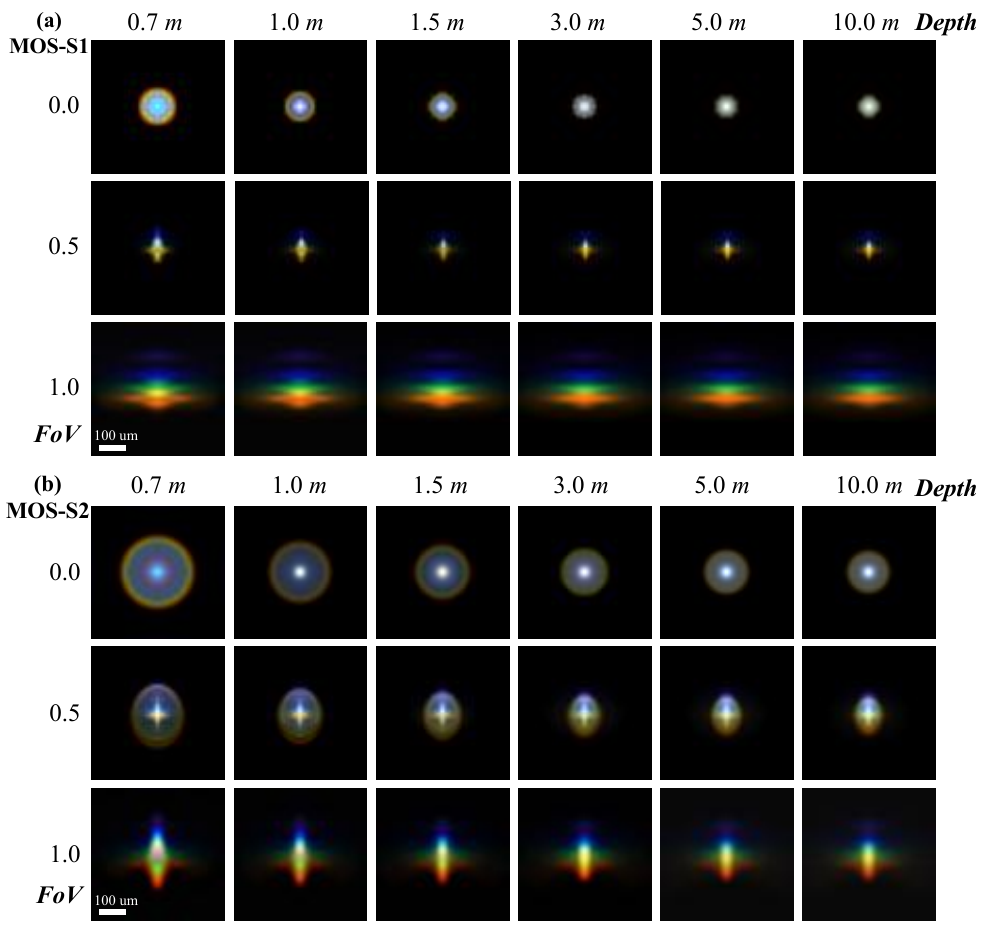}
  \caption{Visualization of PSFs across varying FoVs (normalized diagonal field $fov~ {\in}~ \left [ 0,1 \right ]$) and scene depths (depth $d ~{\in} ~\left [ 0.7~m,10~m \right ] $ and depth $d~{\ge}~10m$ is regarded as infinity). The PSFs of MOS-S1 and MOS-S2 are both spatially non-uniform and exhibit depth-dependent variations. The MOS-S1 has a focal length of $20.0~mm$, an F-number of $5.0$, and a FoV of $42.6^{\circ}$. The MOS-S2 has a focal length of $24.7~mm$, an F-number of $3.5$, and a FoV of $34.0^{\circ}$. \JQ{The size of each PSF is $41{\times}41 $ at a pixel size of $12.2~{\mu}m$.}}
  \label{fig:PSF simulation}
\end{figure}

\subsection{Implementation Details}
All the experiments of our work are implemented on a single A800 GPU. When training the CAC model, loss is calculated by L1 distance and optimized by the Adam optimizer~\cite{kingma2014adam} with $\beta_1{=}0.9$, $\beta_2{=}0.99$. 
We train the CAC models with an initial learning rate of $2e{-}4$ and a batch size of $8$.
We also use random crop, flip, and rotation for data augmentation and randomly crop $256{\times}256$ image patch for training. 
In addition, we train Omni-Lens-Field for $600,000$ iterations by employing L2 loss to fit the 4D PSFLib of different lenses and apply CosineAnnealing~\cite{loshchilov2016sgdr} learning rate scheduler and AdamW optimizer~\cite{loshchilov2017decoupled} with default parameters. The initial learning rate is set to $1e{-}4$.

\subsection{Depth-aware PSF Simulation}

\begin{table*}[!t]
\centering
\caption{\JQ{Quantitative comparison of image restoration results between the two training schemes.}}
\label{tab:comapre_pipeline}
\renewcommand\arraystretch{1.2}
\huge
\resizebox{0.85\textwidth}{!}{%
\begin{tabular}{|c|c|llllll|llllll|}
\hline
\multirow{3}{*}{Method} & \multirow{3}{*}{Training Scheme} & \multicolumn{6}{c|}{MOS-S1}                                                                                                                                                        & \multicolumn{6}{c|}{MOS-S2}                                                                                                                                                        \\ \cline{3-14} 
                        &                                  & \multicolumn{3}{c|}{NYUv2}                                                                   & \multicolumn{3}{c|}{MiddleBury2014}                                                 & \multicolumn{3}{c|}{NYUv2}                                                                   & \multicolumn{3}{c|}{MiddleBury2014}                                                 \\ \cline{3-14} 
                        &                                  & \multicolumn{1}{c}{PSNR$\uparrow$} & \multicolumn{1}{c}{SSIM$\uparrow$} & \multicolumn{1}{c|}{LPIPS$\downarrow$}          & \multicolumn{1}{c}{PSNR$\uparrow$} & \multicolumn{1}{c}{SSIM$\uparrow$} & \multicolumn{1}{c|}{LPIPS$\downarrow$} & \multicolumn{1}{c}{PSNR$\uparrow$} & \multicolumn{1}{c}{SSIM$\uparrow$} & \multicolumn{1}{c|}{LPIPS$\downarrow$}          & \multicolumn{1}{c}{PSNR$\uparrow$} & \multicolumn{1}{c}{SSIM$\uparrow$} & \multicolumn{1}{c|}{LPIPS$\downarrow$} \\ \hline
\multirow{2}{*}{RB~\cite{lim2017enhanced}}     & Baseline                         & 37.6845                   & 0.9717                    & \multicolumn{1}{l|}{0.0399}          & 35.5927                   & 0.9590                    & 0.0785                      & 30.9015                   & 0.9046                    & \multicolumn{1}{l|}{0.1080}          & 33.2049                   & 0.9213                    & 0.1305                      \\
                        & Depth-aware (Ours)                      & \textbf{38.0825}          & \textbf{0.9733}           & \multicolumn{1}{l|}{\textbf{0.0372}} & \textbf{35.6724}          & \textbf{0.9594}           & \textbf{0.0766}             & \textbf{34.4400}          & \textbf{0.9421}           & \multicolumn{1}{l|}{\textbf{0.0730}} & \textbf{35.2691}          & \textbf{0.9433}           & \textbf{0.1081}             \\ \hline
\multirow{2}{*}{RDG~\cite{Hsu_2024_CVPR}}    & Baseline                         & 37.7239                   & 0.9713                    & \multicolumn{1}{l|}{0.0402}          & 35.0546                   & 0.9576                    & 0.0799                      & 30.8689                   & 0.9040                    & \multicolumn{1}{l|}{0.1061}          & 33.1028                   & 0.9202                    & 0.1349                      \\
                        & Depth-aware (Ours)                      & \textbf{38.3034}          & \textbf{0.9735}           & \multicolumn{1}{l|}{\textbf{0.0370}} & \textbf{35.2324}          & \textbf{0.9580}           & \textbf{0.0777}             & \textbf{33.9388}          & \textbf{0.9373}           & \multicolumn{1}{l|}{\textbf{0.0761}} & \textbf{34.5117}          & \textbf{0.9381}           & \textbf{0.1146}             \\ \hline
\multirow{2}{*}{PSA~\cite{zhou2023srformer}}    & Baseline                         & 38.1562                   & 0.9737                    & \multicolumn{1}{l|}{0.0378}          & 35.1990                   & 0.9594                    & 0.0784                      & 31.3176                   & 0.9080                    & \multicolumn{1}{l|}{0.1047}          & 33.3769                   & 0.9235                    & 0.1313                      \\
                        & Depth-aware (Ours)                      & \textbf{38.7151}          & \textbf{0.9753}           & \multicolumn{1}{l|}{\textbf{0.0354}} & \textbf{35.6695}          & \textbf{0.9598}           & \textbf{0.0780}             & \textbf{34.7431}          & \textbf{0.9445}           & \multicolumn{1}{l|}{\textbf{0.0704}} & \textbf{35.4583}          & \textbf{0.9442}           & \textbf{0.1085}             \\ \hline
\multirow{2}{*}{RSTB~\cite{liang2021swinir}}   & Baseline                         & 38.1762                   & 0.9739                    & \multicolumn{1}{l|}{0.0369}          & 35.2696                   & 0.9606                    & 0.0761                      & 31.3283                   & 0.9086                    & \multicolumn{1}{l|}{0.1042}          & 33.3981                   & 0.9236                    & 0.1312                      \\
                        & Depth-aware (Ours)                     & \textbf{38.8280}          & \textbf{0.9758}           & \multicolumn{1}{l|}{\textbf{0.0342}} & \textbf{35.8148}          & \textbf{0.9616}           & \textbf{0.0736}             & \textbf{34.8523}          & \textbf{0.9450}           & \multicolumn{1}{l|}{\textbf{0.0693}} & \textbf{35.3416}          & \textbf{0.9446}           & \textbf{0.1055}             \\ \hline
\end{tabular}%
}
\end{table*}

\begin{figure*}[!t]
  \centering
  \includegraphics[width=1.0\linewidth]{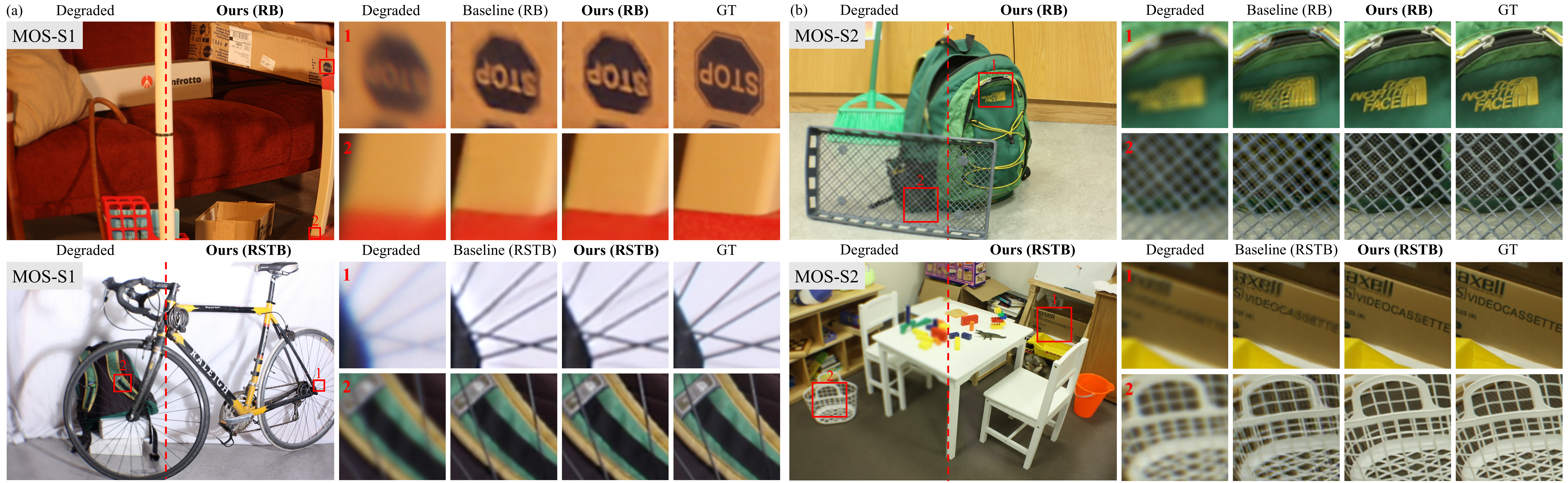}
  \caption{Comparison of the qualitative results of two training schemes on the simulated dataset. (a) and (b) represent the comparison of restoration results on the MOS-S1 and MOS-S2 datasets respectively. We zoom in on image patches at different depths to show details for easier observation. The model trained by the DA$^{2}$T scheme can restore objects at different depths with more texture details and fewer artifacts.} 
  \label{fig:pipeline compare}
\end{figure*}

By conducting ray tracing with varying object planes, the depth-aware spatially-varying PSF of the MOS can be calculated. 
As shown in Fig.~\ref{fig:PSF simulation}, we show examples of PSFs for MOS-S1 and MOS-S2 at different FoVs and depths.
The simulation results intuitively show that the PSFs vary with depth and are spatially non-uniform, which is caused by the optical aberrations introduced by the imperfections of the MOS. 
It is worth noting that the PSF of MOS-S2 changes more significantly with depth compared to that of MOS-S1.

\subsection{\JQ{Results of Monocular Depth Estimation}}
\JQ{We present the depth estimation results, where the GT images and aberration images are predicted using the same metric MDE model~\cite{piccinelli2024unidepth}.}

\JQ{As shown in Table~\ref{tab:unidepth_gt_aber}, quantitative experimental results indicate that the depth predicted from clear images is not always superior to that predicted from aberration images. 
On the Middlebury2014 dataset~\cite{scharstein2014high}, the depth predicted from aberration images can even outperform that from clear images. 
Therefore, aberration does not consistently have a negative impact on depth prediction; in some cases, it can implicitly encode depth information, thereby enhancing the accuracy of the predictions~\cite{chang2019deep,ikoma2021depth,tan20213d,liu2022investigating}.
}

\JQ{It is worth noting that the distortion introduced by the optical system does not impact the overall image quality. }
\JQ{
Since this paper focuses on aberration correction, we ensure pixel alignment between the GT image and the aberration image by applying the same optical system distortion to both during simulation. 
Consequently, the GT images vary between different optical systems, which explains the discrepancies in the quantitative metrics of the depth maps predicted by the same MDE model for GT images from two different optical systems.
}
\JQ{Compared to the depth predictions from GT images, the depth predictions from aberration images still provide reliable depth information.}
\JQ{The qualitative results are shown in the appendix.}

\begin{table}[!t]
\centering
\caption{Quantitative comparison of monocular depth estimation using the same model on GT images and aberration images.}
\label{tab:unidepth_gt_aber}

\renewcommand\arraystretch{1.2}
\huge
\resizebox{1.0\columnwidth}{!}{%
\begin{tabular}{|c|c|c|cccccc|}
\hline
\multirow{3}{*}{Lens name} & \multirow{3}{*}{Method}   & \multirow{3}{*}{Image} & \multicolumn{6}{c|}{Dataset}                                                                                              \\ \cline{4-9} 
                           &                           &                           & \multicolumn{3}{c|}{NYUv2}                                            & \multicolumn{3}{c|}{MiddleBury2014}              \\ \cline{4-9} 
                           &                           &                           & AbsRel~$\downarrow$         & RMSE~$\downarrow$           & \multicolumn{1}{c|}{$\delta>1.25\uparrow$}             & AbsRel~$\downarrow$         & RMSE~$\downarrow$           & $\delta>1.25\uparrow$             \\ \hline
\multirow{2}{*}{MOS-S1}    & \multirow{4}{*}{Unidepth~\cite{piccinelli2024unidepth}} & GT                        & \textbf{0.082} & \textbf{0.328} & \multicolumn{1}{c|}{\textbf{0.955}} & 0.429          & \textbf{1.143} & \textbf{0.321} \\
                           &                           & Aberration                & 0.096          & 0.366          & \multicolumn{1}{c|}{0.934}          & \textbf{0.425} & 1.179          & 0.273          \\ \cline{1-1} \cline{3-9} 
\multirow{2}{*}{MOS-S2}    &                           & GT                        & \textbf{0.076} & \textbf{0.311} & \multicolumn{1}{c|}{\textbf{0.961}} & 0.477          & 1.292          & 0.229          \\
                           &                           & Aberration                & 0.097          & 0.369          & \multicolumn{1}{c|}{0.930}          & \textbf{0.442} & \textbf{1.254} & \textbf{0.300} \\ \hline
\end{tabular}%
}
\end{table}

\subsection{Results of Synthetic Images Restoration}

For synthetic datasets with ground truth, PSNR and SSIM~\cite{wang2004image} are widely employed to evaluate the fidelity of restored images. 
In addition, we use LPIPS~\cite{zhang2018unreasonable} to evaluate the perceptual quality.

\begin{table*}[!t]
\centering
\caption{\JQ{Quantitative image restoration results of our depth-aware mechanism under the same DA$^{2}$T scheme.}}

\label{tab:depth-aware mechanism}
\renewcommand\arraystretch{1.2}
\huge
\resizebox{0.85\textwidth}{!}{%
\begin{tabular}{|c|c|llllll|llllll|}
\hline
\multirow{3}{*}{Method} & \multirow{3}{*}{\begin{tabular}[c]{@{}c@{}}Depth-aware\\ Mechanism\end{tabular}} & \multicolumn{6}{c|}{MOS-S1}                                                                                                                                                        & \multicolumn{6}{c|}{MOS-S2}                                                                                                                                                        \\ \cline{3-14} 
                        &                                        & \multicolumn{3}{c|}{NYUv2}                                                                   & \multicolumn{3}{c|}{MiddleBury2014}                                                 & \multicolumn{3}{c|}{NYUv2}                                                                   & \multicolumn{3}{c|}{MiddleBury2014}                                                 \\ \cline{3-14} 
                        &                                        & \multicolumn{1}{c}{PSNR$\uparrow$} & \multicolumn{1}{c}{SSIM$\uparrow$} & \multicolumn{1}{c|}{LPIPS$\downarrow$}          & \multicolumn{1}{c}{PSNR$\uparrow$} & \multicolumn{1}{c}{SSIM$\uparrow$} & \multicolumn{1}{c|}{LPIPS$\downarrow$} & \multicolumn{1}{c}{PSNR$\uparrow$} & \multicolumn{1}{c}{SSIM$\uparrow$} & \multicolumn{1}{c|}{LPIPS$\downarrow$}          & \multicolumn{1}{c}{PSNR$\uparrow$} & \multicolumn{1}{c}{SSIM$\uparrow$} & \multicolumn{1}{c|}{LPIPS$\downarrow$} \\ \hline
\multirow{2}{*}{RB~\cite{lim2017enhanced}}     & \ding{55}                                      & 38.0825                   & 0.9733                    & \multicolumn{1}{l|}{0.0372}          & 35.6724                   & 0.9594                    & 0.0766                      & 34.4400                   & 0.9421                    & \multicolumn{1}{l|}{0.0730}          & 35.2591                   & 0.9433                    & 0.1081                      \\
                        & \ding{51}                                     & \textbf{38.2605}          & \textbf{0.9741}           & \multicolumn{1}{l|}{\textbf{0.0352}} & \textbf{35.8506}          & \textbf{0.9597}           & \textbf{0.0750}             & \textbf{34.6096}          & \textbf{0.9449}           & \multicolumn{1}{l|}{\textbf{0.0698}} & \textbf{35.5162}          & \textbf{0.9445}           & \textbf{0.1021}             \\ \hline
\multirow{2}{*}{RDG~\cite{Hsu_2024_CVPR}}    & \ding{55}                                       & 38.3034                   & 0.9735                    & \multicolumn{1}{l|}{0.0370}          & 35.2324                   & 0.9580                    & 0.0777                      & 33.9388                   & 0.9373                    & \multicolumn{1}{l|}{0.0761}          & 34.5117                   & 0.9381                    & 0.1146                      \\
                        & \ding{51}                                      & \textbf{38.6161}          & \textbf{0.9736}           & \multicolumn{1}{l|}{\textbf{0.0355}} & \textbf{35.4396}          & \textbf{0.9592}           & \textbf{0.0759}             & \textbf{34.4879}          & \textbf{0.9423}           & \multicolumn{1}{l|}{\textbf{0.0724}} & \textbf{35.3926}          & \textbf{0.9435}           & \textbf{0.1071}             \\ \hline
\multirow{2}{*}{PSA~\cite{zhou2023srformer}}    & \ding{55}                                       & 38.7151                   & 0.9753                    & \multicolumn{1}{l|}{0.0354}          & 35.6695                   & 0.9598                    & 0.0780                      & 34.7431                   & 0.9445                    & \multicolumn{1}{l|}{0.0704}          & 35.4583                   & 0.9442                    & 0.1085                      \\
                        & \ding{51}                                      & \textbf{38.9214}          & \textbf{0.9759}           & \multicolumn{1}{l|}{\textbf{0.0344}} & \textbf{35.8397}          & \textbf{0.9608}           & \textbf{0.0764}             & \textbf{34.9810}          & \textbf{0.9464}           & \multicolumn{1}{l|}{\textbf{0.0688}} & \textbf{35.5914}          & \textbf{0.9464}           & \textbf{0.1048}             \\ \hline
\multirow{2}{*}{RSTB~\cite{liang2021swinir}}   & \ding{55}                                       & 38.8280                   & 0.9758                    & \multicolumn{1}{l|}{0.0342}          & 35.8148                   & 0.9616                    & 0.0736                      & 34.8523                   & 0.9450                    & \multicolumn{1}{l|}{0.0693}          & 35.3416                   & 0.9446                    & 0.1055                      \\
                        & \ding{51}                                      & \textbf{39.1597}          & \textbf{0.9763}           & \multicolumn{1}{l|}{\textbf{0.0334}} & \textbf{36.2204}          & \textbf{0.9624}           & \textbf{0.0735}             & \textbf{35.0233}          & \textbf{0.9466}           & \multicolumn{1}{l|}{\textbf{0.0679}} & \textbf{35.5722}          & \textbf{0.9466}           & \textbf{0.1028}             \\ \hline
\end{tabular}%
}
\end{table*}

\begin{figure*}[!t]

  \centering
  \includegraphics[width=0.85\linewidth]{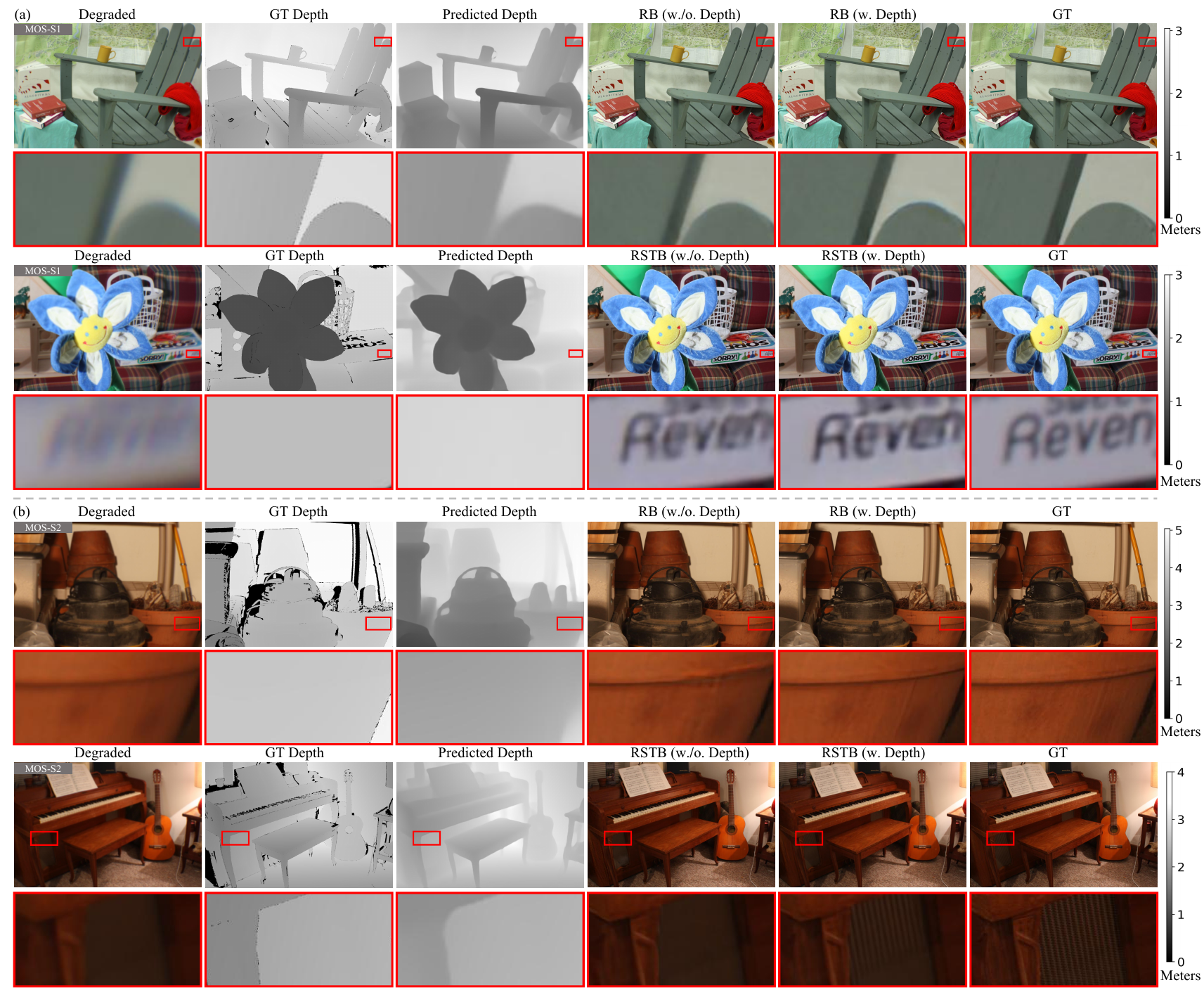}

  \caption{Qualitative results of depth-aware mechanism under the same DA$^{2}$T scheme on the simulated dataset. (a) and (b) display the comparative restoration results on the MOS-S1 and MOS-S2 datasets, respectively. The black regions in the GT depth represent missing values. The depth of the degraded image is estimated using the UniDepth model~\cite{piccinelli2024unidepth}. We compare results with and without the depth-aware mechanism block. The depth-aware mechanism method produces richer and more realistic image details. The reconstruction differences are highlighted by using red boxes.}
  \label{fig:depth_mechanism}

\end{figure*}

\begin{figure}[!t]

  \centering
  \includegraphics[width=1.0\linewidth]{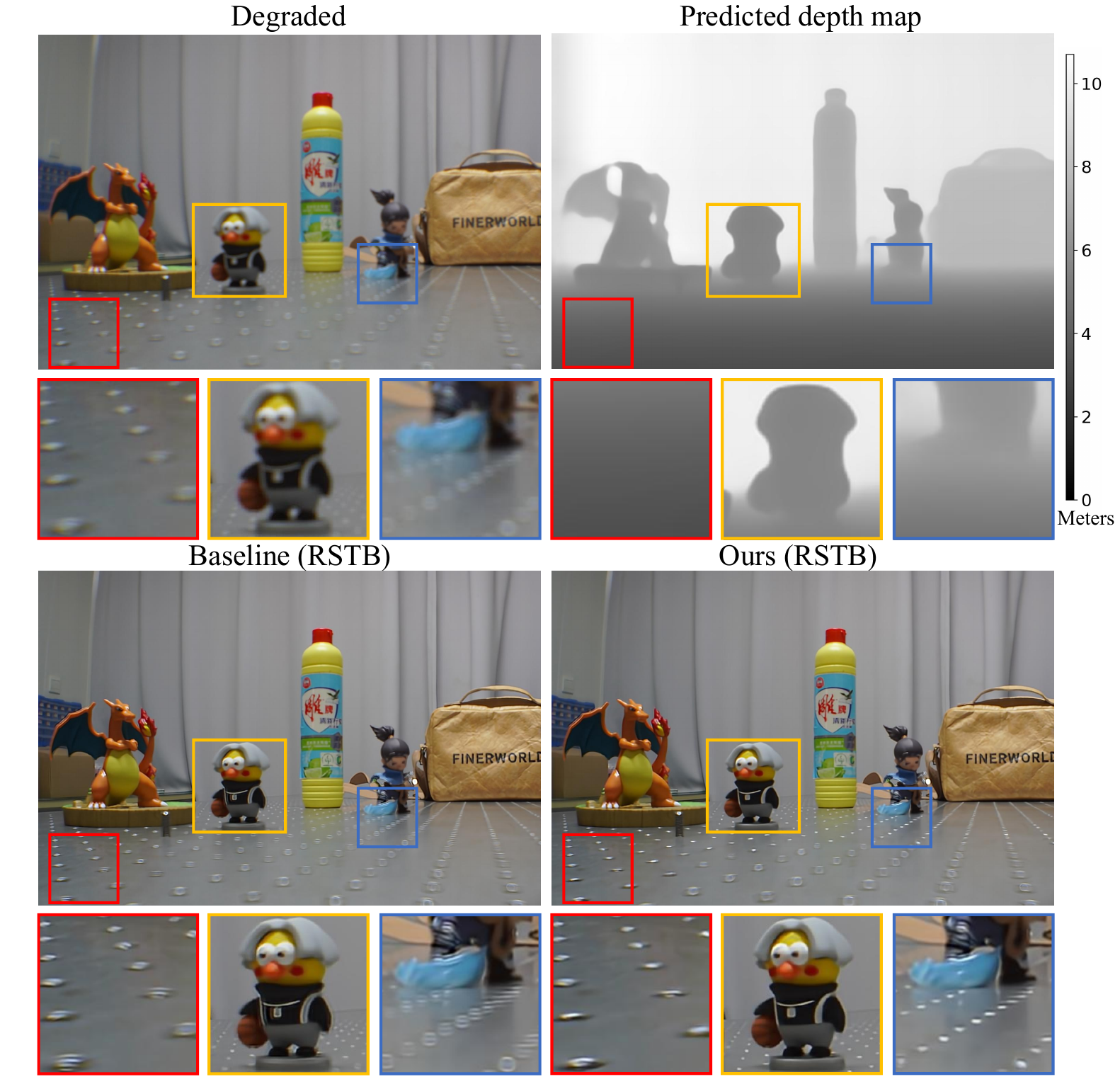}

  \caption{Qualitative results on the real-world \JQ{indoor} dataset. We constructed a scene with diverse objects positioned at varying depths within an optical laboratory under constant lighting conditions to capture a real-world \JQ{indoor} dataset. The real-world \JQ{indoor} dataset is captured using MOS-S1, which has also been employed to evaluate pre-trained CAC models. The DA$^{2}$T scheme and depth-aware mechanism effectively restore objects at different depths.}
  \label{fig:real world}

\end{figure}

\noindent \textbf{Comparison of recovery results of different training schemes.} 
For comparison purposes, we first simulate a depth-unaware training dataset as our baseline, which is obtained by convolving the PSF map of the same scene depth.
To verify the generalization of our proposed DA$^{2}$T scheme, we select the main components of four representative state-of-the-art SR models, including a CNN-based Module, \textit{i.e.}, Residual Block (RB) in EDSR~\cite{lim2017enhanced} and three transformer-based Modules including Residual Swin Transformer Block (RSTB) in SwinIR~\cite{liang2021swinir}, Permuted Self-Attention (PSA) in SRformer~\cite{zhou2023srformer}, and Residual Deep-feature-extraction Group (RDG) in DRCT~\cite{Hsu_2024_CVPR}) for our experiments. To ensure a fair comparison, the experimental settings of the two training schemes are the same, including model parameters, initial learning rate, decay strategy, and number of training epochs. 

As shown in Table~\ref{tab:comapre_pipeline}, the numerical results show that compared with the baseline, the model trained by the DA$^{2}$T scheme has a stable improvement on four different models on two MOS. 
This proves that the scheme does not rely on a certain model and generalizes well to various architectures with consistent gains. 
On the MOS-S1 simulation data, the model trained using the DAT scheme outperforms the baseline by a margin of $0.398dB{\sim}0.6518dB$ on the NYUv2 dataset, and by $0.0797dB{\sim}0.5452dB$ on the Middlebury2014 dataset. 
On the MOS-S2 simulation dataset, the DA$^{2}$T-trained model demonstrates superior performance, surpassing the baseline by $3.0699dB{\sim}3.5300dB$ and $1.4089dB{\sim}2.0814dB$. 
It can be observed that, compared with MOS-S1, the performance gap between the models trained by the two training strategies is more significant in MOS-S2. 
This is because the PSFs in the MOS-S2 dataset change more dramatically with the scene depth. 
These experimental results indicate that the model trained using the DA$^{2}$T scheme effectively restores degraded scenes with varying depths. 

In Fig.~\ref{fig:pipeline compare}, we present qualitative results of the recovery images. 
To verify that the DA$^{2}$T scheme can flexibly apply and generalize to different network structures, we select the recovery results of the CNN-based RB module~\cite{lim2017enhanced} and the transformer-based RSTB module~\cite{liang2021swinir} for display.
Compared to the baseline, the CAC model trained by the DA$^{2}$T scheme delivers moderately clear, aberration-free images that can restore the clarity of objects at different depths. 

\begin{figure*}[!t]
  \centering
  \includegraphics[width=1.0\linewidth]{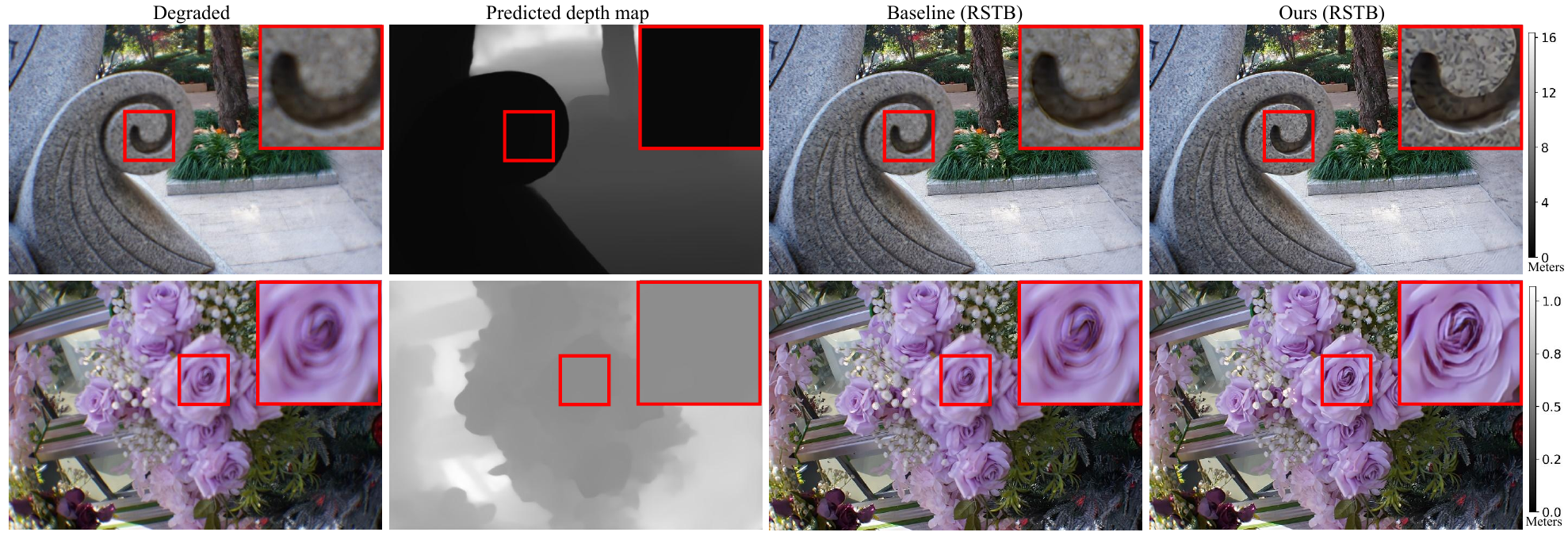}
  \caption{Qualitative results on the real-world outdoor dataset. The real-world outdoor dataset is captured using MOS-S1, which has also been employed to evaluate pre-trained CAC models. The proposed training scheme and depth-aware mechanism effectively restore outdoor scenes with dynamic depth variations.}
  \label{fig:real_world_outdoor}
\end{figure*}

\begin{figure*}[!t]
  \centering
  \includegraphics[width=0.95\linewidth]{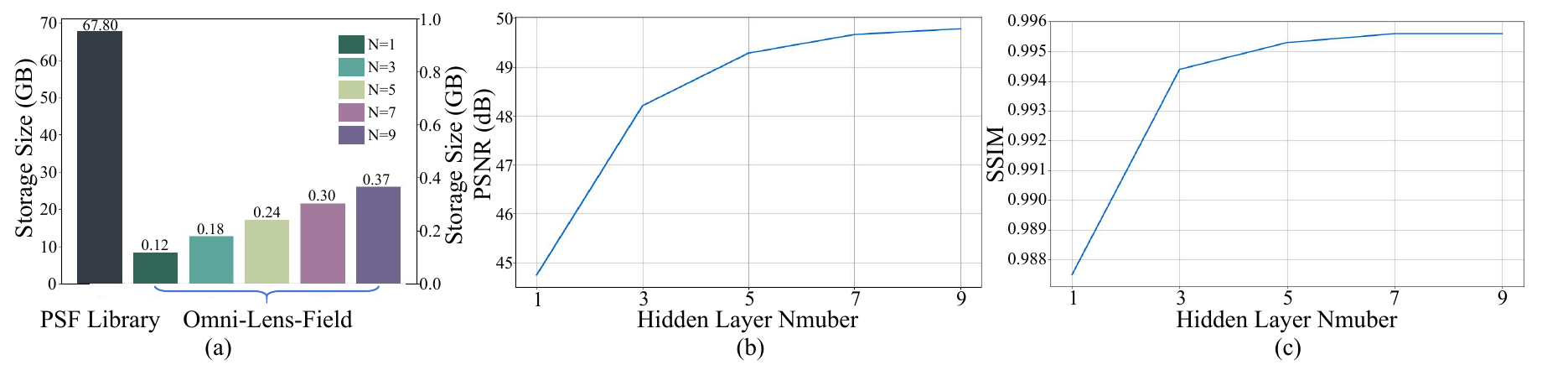}
  \caption{Comparison results of different hidden layer number settings. (a) Storage size of Omni-Lens-Field with different hidden layer number settings and 4D PSFLib with four lenses. (b) and (c) are the quantitative metrics of Omni-Lens-Field of different hidden layer settings on the test set.}
  \label{fig:mlp_N}
\end{figure*}

\noindent \textbf{Restoration experiment of depth-aware mechanism.} 
Similar to the experiment in the previous section, we sequentially replaced the RACM in DACN with four representative state-of-the-art super-resolution models, to investigate the recovery capacity of our proposed depth-aware mechanisms (RDICAB and D$^{2}$CB). 
The quantitative results are shown in Table~\ref{tab:depth-aware mechanism}. Across the four different network architectures, the designed depth-aware mechanisms exhibit stable improvements in the evaluation metrics. 
On the MOS-S1 dataset, the two proposed plug-and-play blocks, RDICAB and D$^{2}$CB, lead to PSNR improvements ranging from $0.1780dB$ to $0.3317dB$. 
On the MOS-S2 dataset, the PSNR gains brought by these two depth-aware mechanism blocks range from $0.1696dB$ to $0.8809dB$. 

As shown in Fig.~\ref{fig:depth_mechanism}, the method equipped with the depth-aware mechanism contributes to a superior visual reconstruction result, faithfully recreating details closer to the ground truth and producing aberration-free images. 
Overall, our proposed two plug-and-play blocks (RDICAB and D$^{2}$CB) demonstrate consistent and pronounced improvements in both quantitative metrics and visual quality across various architectures, showcasing the powerful performance and robust generalization capabilities of the depth-aware mechanism for aberration correction.

\subsection{Results of Real-world Images Restoration}

\textbf{\JQ{Indoor Scene Evaluation.}}
The qualitative results of a real-world scene are displayed in Fig.~\ref{fig:real world}, where we set up a scene in an optical laboratory under constant lighting conditions, with objects placed at varying depths. 
When restoring real-world data, we utilize RSTB as the RACM component in our proposed DACN, as it demonstrates the best performance on the simulated dataset. 
The DA$^{2}$T model equipped with the depth-aware mechanism demonstrates superior visual results compared to the baseline (non-DA$^{2}$T model without the depth-aware mechanism). 
By zooming in on image patches at different depths, it can be observed that our method successfully restores the texture information and reduces artifacts.

\textbf{\JQ{Outdoor Scene Evaluation.}}
\JQ{The qualitative results of outdoor scenes are displayed in Fig.~\ref{fig:real_world_outdoor}. 
Real-world outdoor scenes, characterized by more dynamic depth variations, present greater challenges for restoration. We observed that the non-DAT model struggles to restore scenes with certain depth ranges, leading to unpleasant artifacts. In contrast, the DAT-trained model, equipped with a depth-aware mechanism, effectively reduces artifacts and delivers visually superior restoration results.}

\subsection{Results of Controllable Depth-of-Field Imaging}

\begin{table}[!t]
    \centering
    \caption{Inference time analysis of Depth Estimation Module, Computational Aberration Correction Modules, and Omni-Lens-Field. The inference time is calculated with the input resolution of $1280{\times}1920$ on a single 3090 GPU. The percentage represents the proportion of time that the Omni-Lens-Field occupies within the entire pipeline.}
    \renewcommand\arraystretch{1.2}
\huge
\resizebox{\columnwidth}{!}{%
\begin{tabular}{c|c|c|c|c|c|c}
\hline
Method                    & \begin{tabular}[c]{@{}c@{}}Inference\\ Time (s)\end{tabular} & Method & \begin{tabular}[c]{@{}c@{}}Inference\\ Time (s)\end{tabular} & Method                                                                      & \begin{tabular}[c]{@{}c@{}}Inference\\ Time (s)\end{tabular} & Percentage \\ \hline
\multirow{4}{*}{UniDepth~\cite{piccinelli2024unidepth}} & \multirow{4}{*}{0.212}                                       & RB~\cite{lim2017enhanced}     & 0.092                                                        & \multirow{4}{*}{\begin{tabular}[c]{@{}c@{}}Omni-Lens-\\ Field\end{tabular}} & \multirow{4}{*}{0.028}                                       & 8.4\%      \\
                          &                                                              & RDG~\cite{Hsu_2024_CVPR}    & 0.657                                                        &                                                                             &                                                              & 3.1\%      \\
                          &                                                              & PSA~\cite{zhou2023srformer}    & 0.369                                                        &                                                                             &                                                              & 4.6\%      \\
                          &                                                              & RSTB~\cite{liang2021swinir}   & 0.752                                                        &                                                                             &                                                              & 2.8\%      \\ \hline
\end{tabular}%
}

    \label{tab:inference time}
\end{table}

\begin{figure*}[!t]
  \centering
  \includegraphics[width=0.85\linewidth]{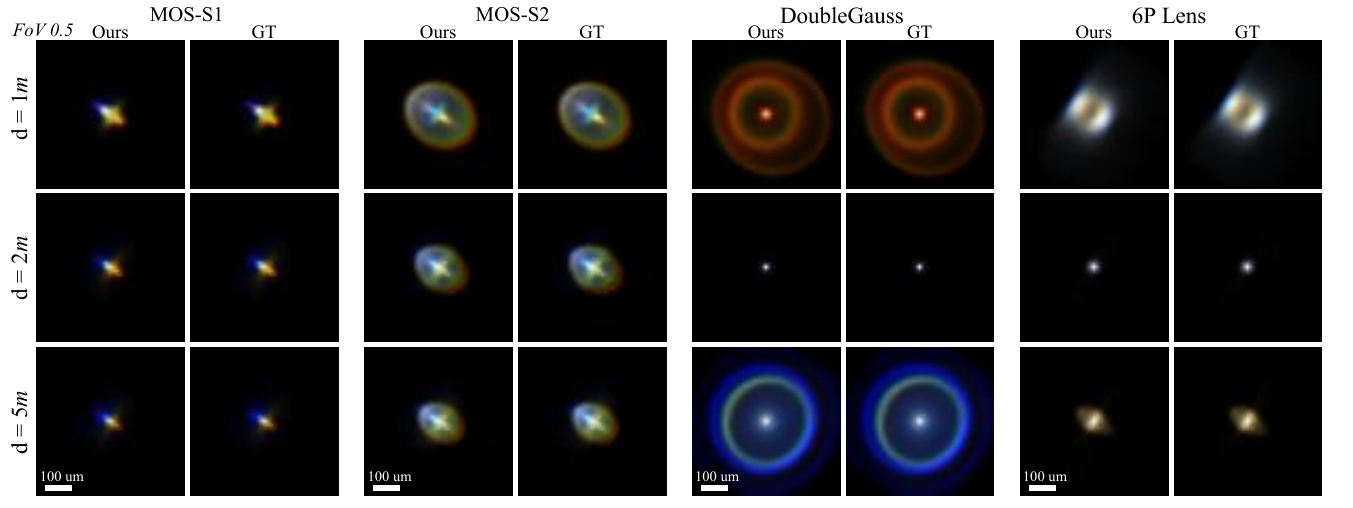}
  \caption{PSF estimation results. We train the Omni-Lens-Field network to represent the 4D PSFLib of four different lenses. The example demonstrates the PSF predicted by our network and ray-tracing PSF at three depths within the same FoV for the four lenses. The PSF produced by our method closely matches the GT. \JQ{The size of each PSF is $41{\times}41 $ at a pixel size of $12.2~{\mu}m$.}}
  \label{fig:mlp_pred}
\end{figure*}

\begin{figure*}[!t]
  \centering
  \includegraphics[width=0.9\linewidth]{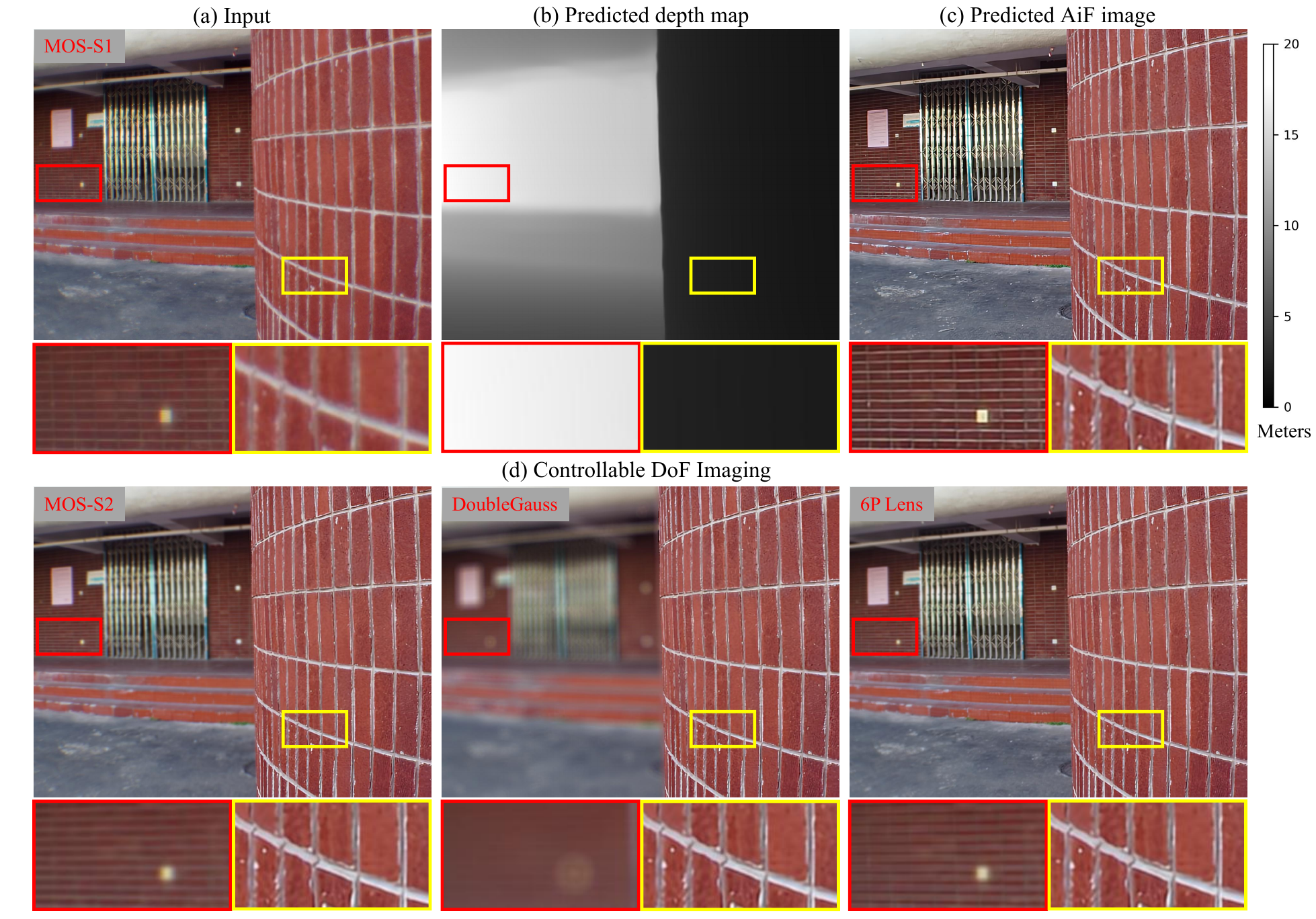}
  \caption{Qualitative results of controllable DoF imaging on the real-world outdoor dataset. (a)~Aberration image simulated by MOS-S1. (b)~Predicted depth map. (c)~Predicted AiF image restored by our DACN. (d) Controllable DoF imaging of different lenses. We select the depth threshold to keep the ``wall'' clear and create the bokeh effects of three lenses, distinct from MOS-S1, beyond the threshold range.}
  \label{fig:real world dof}
\end{figure*}
In our experiment, we aim to use the Omni-Lens-Field model to characterize the 4D PSFLib of four different lenses: the MOS-S1, MOS-S2, a Double Gauss lens, and a 6P lens automatically designed by OptiFusion~\cite{gao2024global}. 
In Fig.~\ref{fig:mlp_N}, we explore the selection of the number of hidden layers $N$, to find the balance between storage space and performance. 
Based on these results, we set $N{=}5$, which allows for precise PSF prediction within a relatively small storage size.
Compared to the original 4D PSFLib of four lenses, which requires $67.8GB$ of storage space, our Omni-Lens-Field model achieves the same representational capacity while occupying only $0.24GB$ of storage. 
This represents a $282.5$-fold reduction in the storage space requirements.
Additionally, the performance of our method, as measured by PSNR and SSIM on the validation set, reaches $49.2855dB$ and $0.9953$ respectively, which confirms that our method is storage-efficient and achieves high-quality prediction with little storage consumption.
\JQ{The time consumption of the model should also be taken into account. As shown in Table~\ref{tab:inference time}, the Omni-Lens-Field accounts for at most 8.4\% of the inference time in the entire pipeline, which is within an acceptable range.}

In Fig.\ref{fig:mlp_pred}, we present the estimated PSF results of four lenses at depths of $1m$, $2m$, and $5m$ with a FoV of $0.5$. 
At each depth, our predictions closely match the ray-tracing PSFs.
The detailed optical parameters of these four lenses and more prediction results are shown in the Appendix.

As depicted in Eq.~\eqref{eq:PSF MAP}, the depth-aware PSF map of the selected lens can be inferred by inputting the predicted depth value $d$ of the scene and the position $(h, w)$ of the corresponding image into the Omni-Lens-Field model.

For a degraded image captured by the MOS, our AiF aberration correction method can recover an AiF aberration-free image that achieves wide DoF imaging. 
Based on the restored AiF image, by leveraging the predicted depth map and the depth-aware PSF map inferred by Omni-Lens-Field, an image can be obtained where the main subject remains distinctly visible, while the background is given the bokeh effect of any desired lens, thereby achieving a shallow DoF effect of different lenses, as depicted in Eq.~\eqref{eq:arbitrary dof imaging}. 
Note that both the subject and the range to be kept sharp are user-controlled. 
\JQ{In this way, controllable DoF imaging is realized. 
The real-world outdoor controllable DoF image results are presented in Fig.~\ref{fig:real world dof}.}

\subsection{Ablation Study}
To verify the effectiveness of our two plug-and-play modules, RDICAB and D$^{2}$CB, we use RB~\cite{lim2017enhanced} as our RACM module and conduct ablation experiments on NYUv2 and Middlebury2014 datasets. 
All models are trained in the same experimental environment and settings. 
As proofed in Table~\ref{tab:ablation}, both RDICAB and D$^{2}$CB are beneficial to improving the aberration image recovery ability of the model, with RDICAB achieving an increase of $0.0864dB$ in PSNR, while D$^{2}$CB resulting in an increase of $0.1328dB$.

\begin{table}[!t]
    \captionsetup{font={small}}
    \caption{Ablation study on RDICAB and D$^{2}$CB.}
    \label{tab:ablation}
    \renewcommand\arraystretch{1.1}
\huge
\centering
\resizebox{0.8\columnwidth}{!}{%
\begin{tabular}{|c|c|c|ccc|ccc|}
\hline
\multirow{3}{*}{Method} &
  \multirow{3}{*}{RDICAB} &
  \multirow{3}{*}{D$^{2}$CB} &
  \multicolumn{3}{c|}{\multirow{2}{*}{NYUv2}} &
  \multicolumn{3}{c|}{\multirow{2}{*}{MiddleBury2014}} \\
                    &   &   & \multicolumn{3}{c|}{}     & \multicolumn{3}{c|}{}     \\ \cline{4-9} 
                    &   &   & PSNR$\uparrow$   & SSIM$\uparrow$  & LPIPS$\downarrow$ & PSNR$\uparrow$   & SSIM$\uparrow$  & LPIPS$\downarrow$ \\ \hline
\multirow{4}{*}{RB~\cite{lim2017enhanced}} & \ding{55} & \ding{55} & 34.4400 & 0.9421 & 0.0730 & 35.2591 & 0.9433 & 0.1081 \\
                    & \ding{51} & \ding{55} & 34.5264 & 0.9428 & 0.0723 & 35.2610 & 0.9443 & 0.1052 \\
                    & \ding{55} & \ding{51} & 34.5728 & 0.9434 & 0.0708 & 35.4612 & 0.9445 & 0.1048 \\
 &
  \ding{51} &
  \ding{51} &
  \textbf{34.6096} &
  \textbf{0.9449} &
  \textbf{0.0698} &
  \textbf{35.5162} &
  \textbf{0.9445} &
  \textbf{0.1021} \\ \hline
\end{tabular}%
}
\end{table}

\JQ{
Additionally, we further investigate the assignments of Query ($Q$), Key ($K$), and Value ($V$) in RDICAB.
}
\JQ{
In Table.~\ref{tab:QKV}, the CAC model, which obtains queries from the image branch and keys and values from the depth branch, achieves better aberration image restoration results.
Under both assignment conditions, we ensure that the long skip residual connection in RDICAB uses image features.
}
\JQ{
As shown in Eq.~\ref{eq:corss att}, assigning $Q$ to the image branch while assigning $K$ and $V$ to the depth branch enables the cross-attention mechanism to generate depth information modulated by image features. 
This modulated depth information is then added to the image features and fed into subsequent processing modules.
The added depth information significantly enhances the model's ability to handle depth-varying degradations.
In contrast, reversing this assignment results in the generation of single-modal image features, which fail to provide additional  information to the subsequent modules, thereby limiting the overall performance of the model.
}

\JQ{Finally, to evaluate the impact of depth maps predicted by aberration and GT images on the CAC task, we conduct comparative experiments using each as the depth branch input for the CAC model.}
\JQ{As illustrated in Table~\ref{tab:ablation_gt_aber_depth}, the two different depth branch inputs have minimal impact on the model's aberration correction performance.}
\JQ{This suggests that directly using a pre-trained MDE model to predict depth from aberration images can be effectively applied to downstream tasks, as the predicted depth maps are reliable.}

\begin{table}[!t]
    \captionsetup{font={small}}
    \centering
    \caption{Ablation study on the assignments of Query ($Q$), Key ($K$), and Value ($V$).}
    \renewcommand\arraystretch{1.1}
\huge
\centering
\resizebox{0.8\columnwidth}{!}{%
\begin{tabular}{|c|c|c|c|cll|cll|}
\hline
\multirow{3}{*}{Method} & \multirow{3}{*}{$Q$} & \multirow{3}{*}{$K$} & \multirow{3}{*}{$V$} & \multicolumn{3}{c|}{\multirow{2}{*}{NYUv2}}                                                    & \multicolumn{3}{c|}{\multirow{2}{*}{MiddleBury2014}}                                           \\
                        &                    &                    &                    & \multicolumn{3}{c|}{}                                                                          & \multicolumn{3}{c|}{}                                                                          \\ \cline{5-10} 
                        &                    &                    &                    & PSNR$\uparrow$                                & \multicolumn{1}{c}{SSIM$\uparrow$} & \multicolumn{1}{c|}{LPIPS$\downarrow$} & PSNR$\uparrow$                                & \multicolumn{1}{c}{SSIM$\uparrow$} & \multicolumn{1}{c|}{LPIPS$\downarrow$} \\ \hline
\multirow{2}{*}{RB~\cite{lim2017enhanced}}     & Image              & Depth              & Depth              & \multicolumn{1}{l}{\textbf{34.5264}} & \textbf{0.9428}           & \textbf{0.0723}             & \multicolumn{1}{l}{\textbf{35.2610}} & \textbf{0.9443}           & \textbf{0.1052}             \\ \cline{2-4}
                        & Depth              & Image              & Image              & \multicolumn{1}{l}{34.4449}          & 0.9422                    & 0.0728                      & \multicolumn{1}{l}{35.1272}          & 0.9430                    & 0.1077                      \\ \hline
\end{tabular}%
}
    \label{tab:QKV}
\end{table}

\begin{table}[!t]
\centering
\caption{Ablation study on the impact of depth predicted by GT images and aberration images on the CAC task.}
\label{tab:ablation_gt_aber_depth}
\renewcommand\arraystretch{1.1}
\huge
\centering
\resizebox{0.75\columnwidth}{!}{%
\begin{tabular}{|c|c|cll|cll|}
\hline
\multirow{3}{*}{Method} & \multirow{3}{*}{Image} & \multicolumn{3}{c|}{\multirow{2}{*}{NYUv2}}                                                    & \multicolumn{3}{c|}{\multirow{2}{*}{MiddleBury2014}}                                           \\
                        &                        & \multicolumn{3}{c|}{}                                                                          & \multicolumn{3}{c|}{}                                                                          \\ \cline{3-8} 
                        &                        & PSNR$\uparrow$                                & \multicolumn{1}{c}{SSIM$\uparrow$ } & \multicolumn{1}{c|}{LPIPS$\downarrow$} & PSNR$\uparrow$                                & \multicolumn{1}{c}{SSIM$\uparrow$ } & \multicolumn{1}{c|}{LPIPS$\downarrow$} \\ \hline
\multirow{2}{*}{RB~\cite{lim2017enhanced}}     & GT                     & \multicolumn{1}{l}{\textbf{34.6373}} & 0.9437                    & 0.0701                      & \multicolumn{1}{l}{35.4822}          & \textbf{0.9447}           & 0.1041                      \\
                        & Aberration             & \multicolumn{1}{l}{34.6096}          & \textbf{0.9440}           & \textbf{0.0698}             & \multicolumn{1}{l}{\textbf{35.5162}} & 0.9445                    & \textbf{0.1021}             \\ \hline
\end{tabular}%
}
\end{table}

\section{Conclusion}
This paper introduces a Depth-aware Controllable DoF Imaging (DCDI) framework to achieve single-lens controllable DoF imaging via AiF aberration correction and MDE. 
To address depth-varying optical degradation, we first establish the Depth-aware Aberration MOS (DAMOS) dataset by conducting depth-aware image simulation and propose the Depth-aware Degradation-adaptive Training (DA$^{2}$T) scheme, which empowers the network to learn the optical aberration degradation characteristics across varying depths in an adaptive manner. 
To further enhance the recovery performance, we design two depth-aware mechanisms that leverage depth information to aid CAC in effectively addressing depth-varying optical degradation.
We propose the storage-efficient Omni-Lens-Field model to precisely characterize the 4D PSFLib of different lenses with low storage consumption. 
Experimental results on the simulated and real-world data verify that the DA$^{2}$T scheme and depth-aware mechanism dramatically improve the recovery ability of the CAC model. 
The DA$^{2}$T model equipped with the depth-aware mechanism can recover more realistic details and produce fewer artifacts. 
By utilizing the recovered image, the predicted depth map, and the depth-aware PSF map estimated by Omni-Lens-Field, our proposed DCDI framework attains impressive single-lens controllable DoF imaging.

\textbf{Limitations and future work directions.}
\JQ{At the dataset level, the depth range of the scene that our method can effectively restore is limited to a certain extent by the depth range of the training dataset.
In future research, creating a paired RGB and depth dataset with a wide variety of scene content and a broader depth range will be crucial for enhancing the applicability of our method.}
\JQ{In ray tracing simulation, the influences of the scattering media during the light transmission process are not considered.
This paper proposes the first framework for single-lens controllable DoF imaging, focusing on common application scenarios, where the scattering has little impact on the simulation results. 
However, when generalizing this framework to other extreme scenarios (such as foggy weather), it is necessary to consider the scattering characteristics of the medium.}
In aberration correction, the focus distance has not been fully considered; instead, the focus distance is fixed. 
This is because, even with adjustable focusing cameras, the focus plane is only positioned at a selected depth during capture, rather than across all depths. 
However, it is essential to consider the focus distance to achieve the imaging effect of a true high-end lens with a single lens, which requires simulating a 5D PSFLib. 
This, unfortunately, significantly increases the computational overhead and incurs cost-prohibitive storage requirements of ray tracing. 
In the future, we intend to utilize a 5D PSFLib for controllable DoF imaging with a single lens. 

{\small
\bibliographystyle{IEEEtran}
\bibliography{bib}
}


\appendices
\counterwithin{figure}{section}
\counterwithin{equation}{section}
\counterwithin{table}{section}

\section{Lens Data}
Our experiments include four types of lenses: the MOS-S1, MOS-S2, a Double Gauss lens, and a 6P lens automatically designed by OptiFusion~\cite{gao2024global}.
The data of the four lenses are shown in Table~\ref{tab:MOS-S1}, Table~\ref{tab:MOS-S2}, Table~\ref{tab:Double Gauss} and Table~\ref{tab:6PLENS}, respectively. 
In our aberration correction process, we use the MOS-S1 and MOS-S2 for experiments to verify our DA$^{2}$T strategy and depth-ware mechanism. 
In our arbitrary DoF imaging process, we utilize the Omni-Lens-Field to represent the 4D PSF library of these lenses.

\begin{table}[h]
\centering
\caption{Lens data for the MOS-S1 used in this paper.}
\label{tab:MOS-S1}
\resizebox{\columnwidth}{!}{%
\begin{tabular}{lccccc}
\hline
\multicolumn{1}{c}{Surface} & Radius{[}mm{]} & Thickness{[}mm{]} & Material(n/V) & Semi-diameter{[}mm{]} & Conic \\ \hline
1 (Aper)                   & Infinite & 0.200  &       & 1.995  & 0 \\
2 (Sphere)                 & -37.536  & 8.500  & H-K9L & 2.123  & 0 \\
3 (Sphere)                 & -8.711   & 5.662  &       & 5.594  & 0 \\
4 (Sphere)                 & Infinite & 4.000  &       & 8.654  & 0 \\
5 (Sphere)                 & Infinite & 10.300 &       & 10.244 & 0 \\
\multicolumn{1}{c}{Sensor} &          &        &       & 14.364 & 0 \\ \hline
\end{tabular}%
}
\end{table}

\begin{table}[h]
\centering
\caption{Lens data for the MOS-S1 used in this work.}
\label{tab:MOS-S2}
\resizebox{\columnwidth}{!}{%
\begin{tabular}{lccccc}
\hline
\multicolumn{1}{c}{Surface} & Radius{[}mm{]} & Thickness{[}mm{]} & Material(n/V) & Semi-diameter{[}mm{]} & Conic \\ \hline
1 (Aper)                   & Infinite & 4.040  &            & 3.524  & 0 \\
2 (Sphere)                 & -52.628  & 9.423  & 1.95, 81.6 & 5.729  & 0 \\
3 (Sphere)                 & -17.628  & 25.212 &            & 8.292  & 0 \\
\multicolumn{1}{c}{Sensor} &          &        &            & 12.820 & 0 \\ \hline
\end{tabular}%
}
\end{table}

\begin{table}[h]
\centering
\caption{Lens data for the Double Gauss Lens used in this work.}
\label{tab:Double Gauss}
\resizebox{\columnwidth}{!}{%
\begin{tabular}{lccccc}
\hline
\multicolumn{1}{c}{Surface} & Radius{[}mm{]} & Thickness{[}mm{]} & Material(n/V) & Semi-diameter{[}mm{]} & Conic \\ \hline
1 (Sphere)                 & 3.552    & 1.399   & LAF2 & 2.215 & 0 \\
2 (Sphere)                 & Infinite & 0.00877 &      & 2.215 & 0 \\
3 (Sphere)                 & 1.704    & 0.939   & PSK3 & 1.470 & 0 \\
4 (Sphere)                 & -16.238  & 0.252   & SF1  & 1.470 & 0 \\
5 (Sphere)                 & 0.986    & 0.434   &      & 0.810 & 0 \\
6 (Aper)                   & Infinite & 0.278   &      & 0.800 & 0 \\
7 (Sphere)                 & -1.642   & 0.270   & SF1  & 0.810 & 0 \\
8 (Sphere)                 & 1.803    & 1.159   & LAF2 & 1.025 & 0 \\
9 (Sphere)                 & -2.334   & 0.447   &      & 1.025 & 0 \\
10 (Sphere)                & 1.659    & 1.044   & LAF2 & 0.930 & 0 \\
11 (Sphere)                & 14.411   & 0.813   &      & 0.930 & 0 \\
\multicolumn{1}{c}{Sensor} &          &         &      & 0.178 & 0 \\ \hline
\end{tabular}%
}
\end{table}

\begin{table}[h]
\centering
\caption{Lens data for the 6PLENS used in this work. This lens is automatically designed by OptiFusion~\cite{gao2024global}}
\label{tab:6PLENS}
\resizebox{\columnwidth}{!}{%
\begin{tabular}{lccccc}
\hline
\multicolumn{1}{c}{Surface} & Radius{[}mm{]} & Thickness{[}mm{]} & Material(n/V) & Semi-diameter{[}mm{]} & Conic \\ \hline
1 (Sphere)                 & 17.783    & 6.222  & 1.69, 52.4 & 12.521 & 0 \\
2 (Sphere)                 & 10.636    & 8.903  &            & 8.634  & 0 \\
3 (Aper)                   & Infinite  & 1.415  &            & 8.131  & 0 \\
4 (Sphere)                 & -76.857   & 6.051  & 1.88, 39.1 & 8.834  & 0 \\
5 (Sphere)                 & -52.104   & 2.822  &            & 11.066 & 0 \\
6 (Sphere)                 & -30.941   & 14.063 & 1.56, 73.2 & 12.045 & 0 \\
7 (Sphere)                 & -19.931   & 1.006  &            & 16.382 & 0 \\
8 (Sphere)                 & 60.192    & 13.640 & 1.84, 66.2 & 18.809 & 0 \\
9 (Sphere)                 & -61.137   & 2.352  &            & 18.425 & 0 \\
10 (Sphere)                & -41.997   & 4.005  & 1.66, 18.9 & 17.952 & 0 \\
11 (Sphere)                & 123.939   & 7.829  &            & 17.757 & 0 \\
12 (Sphere)                & 55.935    & 10.403 & 1.94, 52.1 & 18.333 & 0 \\
13 (Sphere)                & -1185.313 & 17.020 &            & 17.364 & 0 \\
\multicolumn{1}{c}{Sensor} &           &        &            & 0.178  & 0 \\ \hline
\end{tabular}%
}
\end{table}

\section{\JQ{Impact of Diffraction Effects on Ray Tracing Simulation}}
\JQ{We explain the reasons why diffraction effects can be ignored in our prototype camera (MOS-S1) from the following three perspectives:
}

\JQ{First, the aperture stop of the MOS-S1 has a diameter of approximately $4mm$, which greatly exceeds the wavelength range of visible light, from $380nm$ to $700nm$.
As a result, since the diameter of the aperture stop is much larger than the wavelength of light, no significant diffraction occurs at the aperture stop.}

\JQ{Besides, MOS-S1 utilizes a spherical lens with a continuously varying surface spherical height. There are no microstructures akin to diffraction elements (such as metasurfaces or diffraction optical elements, DOE), which means that no diffraction effects occur on the optical surface.
}

\JQ{Lastly, following your suggestions, we also calculate the Airy disk of this optical system and compare it with the pixel size.
The Airy disk calculation formula is as follows:
\begin{equation}
    d_{Airy}=2.44\cdot  \lambda \cdot \text{F\#},
    \label{eq:Ariy}
\end{equation}
where \text{F\#} denotes image-space F-number, and $\lambda$ denotes visible light wavelength.
Taking ${\lambda}{=}550nm$ in the formula, the diameter of the Airy disk is approximately $6.7{\mu}m$.
When ray tracing the PSF simulation, the image resolution is set to $1280 {\times} 1920$, and the physical diagonal size of the CMOS sensor is $28.2mm$, so the physical size of each pixel is about $12.2{\mu}m$. 
Therefore, the Airy disk diameter is smaller than the physical size of a pixel and the diffraction effect can be ignored.
}

\JQ{In summary, it is reasonable to ignore the influence of diffraction effects in our prototype camera.}

\section{\JQ{Depth-aware Degradation-adaptive Training pipeline}}

\begin{figure*}[!t]
  \centering
  \includegraphics[width=1.0\linewidth]{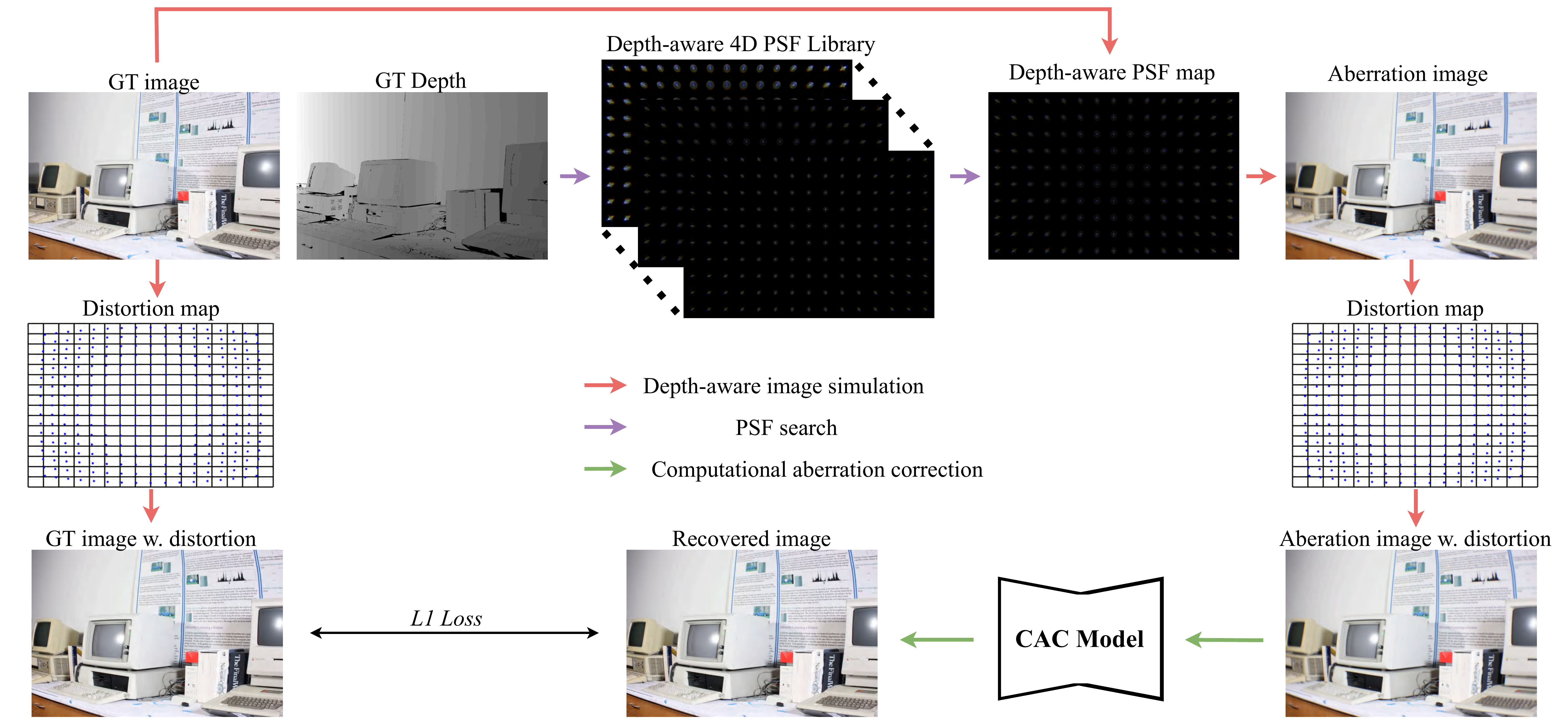}
  
  \caption{Depth-aware Degradation-adaptive Training pipeline. (a) GT image and corresponding depth map are given as the input. (b) based on the depth map, a corresponding PSF is searched in the PSF library for each patch according to the depth value, forming a depth-aware PSF map (purple path). (c) the GT image and the corresponding PSF map are processed using patch-wise convolution to generate the aberration image. Distortion simulation is then both the GT image and the aberration image, which are used as the training data. (pink path). (d) the CAC model takes the aberration image to estimate the recovered image (green path).}
  \label{fig:training scheme}
\end{figure*}

\JQ{In order to embed the depth-varying optical degradation into the network training, the DA$^{2}$T scheme integrates the depth-aware image simulation module into the training process, as illustrated in Fig.~\ref{fig:training scheme}(c). 
GT image and its corresponding depth map are used as input during the overall pipeline (Fig.~\ref{fig:training scheme}(a)).
}

\JQ{
According to the depth map, the corresponding PSF is retrieved from the PSF library for each patch, and a depth-aware PSF map is formed based on the depth value, which is used to generate the aberration image.
The entire training process enables the network to adaptively learn to handle the optical degradation process that varies with depth.
}

\section{More PSF Estimation Results}

\begin{figure*}[p]

  \centering
  \includegraphics[width=0.9\linewidth]{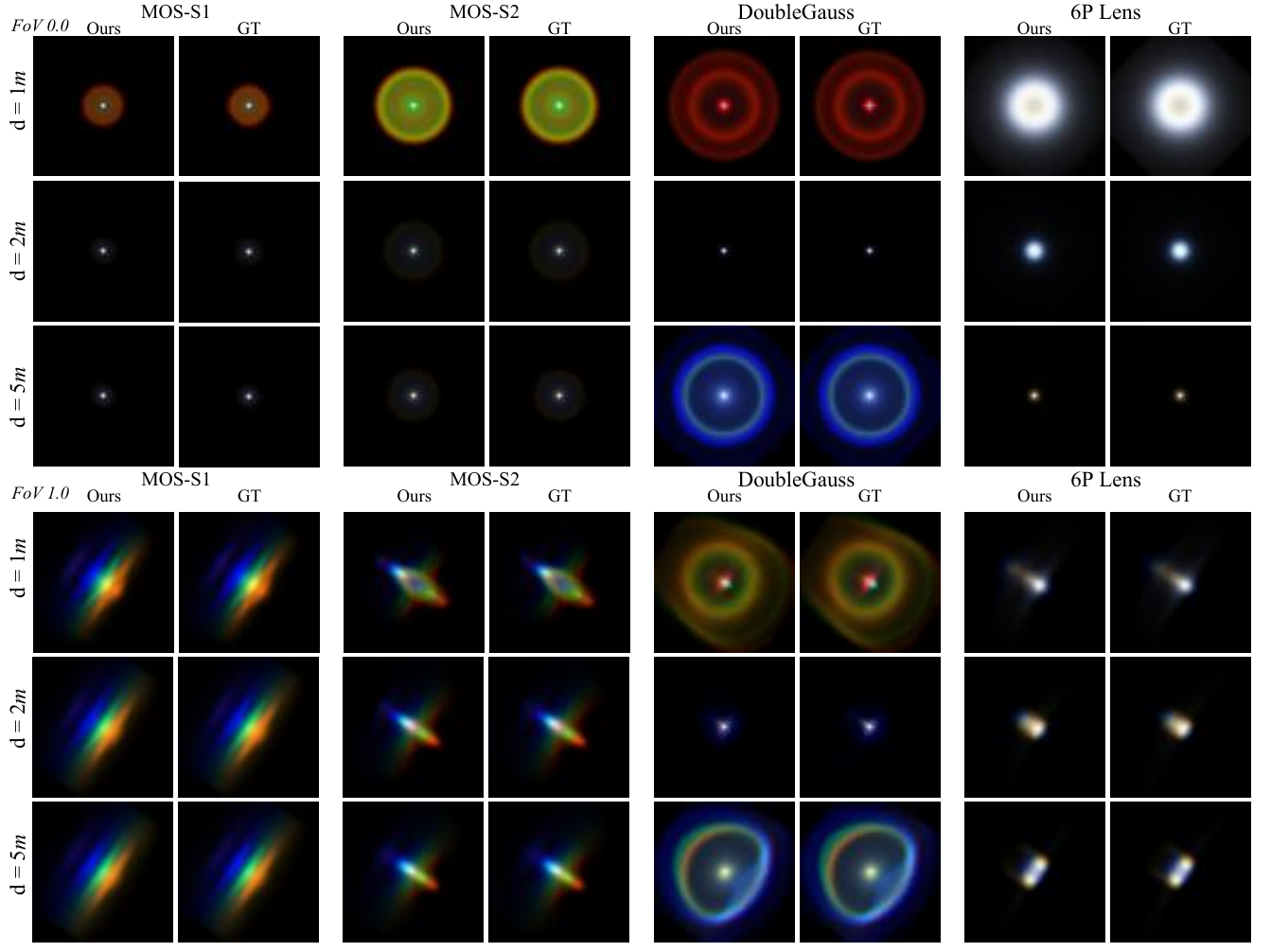}

  \caption{More PSF estimation results. This figure shows the PSF prediction results of our model at two FoVs ($0$ and $1$) and three depths ($1m$, $2m$, $5m$). The PSF predicted by our model is close to GT.}
  \label{fig:morePSFresults}

\end{figure*}

In this section, we show the PSFs of four lenses estimated by the Omni-Lens-Field model at different fields of view and depths, including more results for two fields of view ($0.0$ and $1.0$) and three depths ($1m$, $2m$, and $5m$). As illustrated in Fig.~\ref{fig:morePSFresults}, our Omni-Lens-Field model can predict PSFs at different fields of view and depths close to GT with high quality, which shows that the 4D PSF library of various lenses can be effectively represented by our model. 

\begin{figure*}[p]

  \centering
  \includegraphics[width=0.9\linewidth]{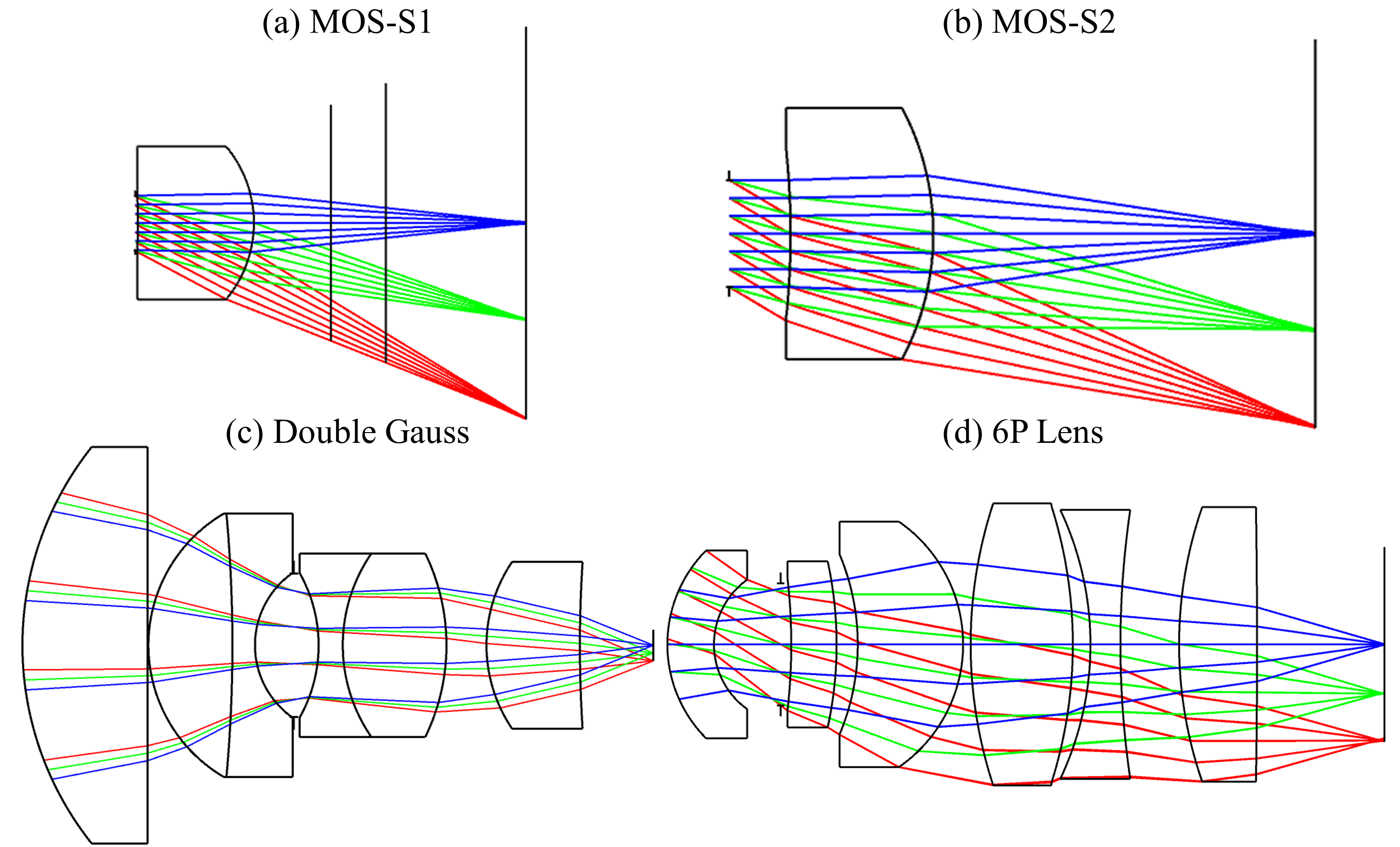}

  \caption{2D views of the four used lenses. (a) The MOS-S1 has a focal length of $20.0mm$, an F-number of $5.0$, and a field of view of $42.6^{\circ}$. (b) The MOS-S2 has a focal length of $24.7mm$, an F-number of $3.5$, and a field of view of $34.0^{\circ}$. (c) The Double Gauss Lens has a focal length of $4.0mm$, an F-number of $1.0$, and a field of view of $2.5^{\circ}$. (d) The 6P Lens has a focal length of $26.44mm$, an F-number of $1.8$, and a field of view of $32.0^{\circ}$.}
  \label{fig:Lens}

\end{figure*}

\section{\JQ{Quantitative Analysis of Different state-of-the-art MDE models on the DAMOS Dataset}}

\JQ{In this section, we compare the depth prediction performance of four state-of-the-art metric MDE models on the DAMOS dataset: ZoeDepth~\cite{bhat2023zoedepth}, Metric3Dv2~\cite{hu2024metric3d}, DepthAnythingV2~\cite{yang2024depthv2}, and Unidepth~\cite{piccinelli2024unidepth}.}

\JQ{To verify the robustness of these metric depth estimation methods, we evaluate on the NYUv2~\cite{silberman2012indoor} and Middlebury2014~\cite{scharstein2014high} datasets of two lenses.}
\JQ{Following previous studies~\cite{bhat2023zoedepth,hu2024metric3d,piccinelli2024unidepth}, we use metrics such as absolute relative error (AbsRel), accuracy under threshold ($\delta_i < 1.25^i, \, i = 1, 2, 3$), root mean squared error (RMSE), root mean squared error in log space ($\text{RMSE}_{\log}$).}

\JQ{Table~\ref{tab:mde_compare} clearly shows Unidepth consistently ranks in the top two across all datasets, demonstrating strong zero-shot generalization capability.
All of the above models use their best-performing versions as presented in the paper.}

\JQ{Based on the above quantitative results, we choose UniDepth as the model to obtain the metric depth information of the scene.}

\begin{table*}[!t]
\centering
\caption{Quantitative comparison of monocular depth estimation on aberration images using different models. ``*'' means that the corresponding camera intrinsics are required for inference. ``\textdagger'' means using its metric depth estimation model.}
\label{tab:mde_compare}
\renewcommand\arraystretch{1.2}
\huge
\resizebox{\textwidth}{!}{
\begin{tabular}{|c|c|cccccc|cccccc|}
\hline
\multirow{2}{*}{Lens name} & \multirow{2}{*}{Method} & \multicolumn{6}{c|}{NYUv2}                                                                                                           & \multicolumn{6}{c|}{Middlebury2014}                                                                 \\ \cline{3-14} 
                           &                         & $\delta>1.25\uparrow$                       & $\delta>1.25^2\uparrow$                       & $\delta>1.25^3\uparrow$                       & AbsRel~$\downarrow$        & RMSE~$\downarrow$          & RMSE$_{log}\downarrow$         & $\delta>1.25\uparrow$            & $\delta>1.25^2\uparrow$            & $\delta>1.25^3\uparrow$            & AbsRel~$\downarrow$        & RMSE~$\downarrow$          & RMSE$_{log}\downarrow$         \\ \hline
\multirow{4}{*}{MOS-S1}    & ZoeDepth~\cite{bhat2023zoedepth}                & \underline{0.886}    & \underline{0.974}    & \underline{0.991}    & \underline{0.113}    & \underline{0.443}    & \underline{0.156}    & \textbf{0.377} & \underline{0.609}    & 0.746          & \textbf{0.335} & \textbf{1.128}    & \textbf{0.457} \\
                           & Metric3Dv2$^{*}$~\cite{hu2024metric3d}                & 0.759          & 0.946          & 0.978          & 0.156          & 0.624          & 0.229          & 0.145          & 0.375          & \underline{0.764}    & 0.828          & 2.106          & 0.578          \\
                           & DepthAnythingv2$^{\dagger}$~\cite{yang2024depthv2}         & 0.791          & 0.960          & 0.984          & 0.167          & 0.621          & 0.201          & 0.130          & 0.283          & 0.612          & 0.465          & 1.345          & 0.678          \\
                           & UniDepth~\cite{piccinelli2024unidepth}                & \textbf{0.934} & \textbf{0.987} & \textbf{0.995} & \textbf{0.096} & \textbf{0.366} & \textbf{0.130} & \underline{0.273}    & \textbf{0.630} & \textbf{0.840} & \underline{0.425}    & \underline{1.179} & \underline{0.465}    \\ \hline
\multirow{4}{*}{MOS-S2}    & ZoeDepth~\cite{bhat2023zoedepth}                & \underline{0.868}    & \underline{0.971}    & \underline{0.991}    & \underline{0.122}    & \underline{0.466}    & \underline{0.166}    & \textbf{0.382} & \textbf{0.604} & \underline{0.751}    & \textbf{0.333} & \textbf{1.117} & \textbf{0.455} \\
                           & Metric3Dv2$^{*}$~\cite{hu2024metric3d}                & 0.788          & 0.949          & 0.976          & 0.144          & 0.581          & 0.223          & 0.089          & 0.313          & 0.700          & 0.928          & 2.381          & 0.625          \\
                           & DepthAnythingv2$^{\dagger}$~\cite{yang2024depthv2}         & 0.712          & 0.948          & 0.983          & 0.204          & 0.671          & 0.219          & 0.122          & 0.306          & 0.636          & 0.464          & 1.338          & 0.682          \\
                           & UniDepth~\cite{piccinelli2024unidepth}                & \textbf{0.930} & \textbf{0.986} & \textbf{0.995} & \textbf{0.097} & \textbf{0.369} & \textbf{0.130} & \underline{0.300}    & \underline{0.583}    & \textbf{0.779} & \underline{0.442}    & \underline{1.254}    & \underline{0.468}    \\ \hline
\end{tabular}%
}
\end{table*}

\section{\JQ{Qualitative Analysis of Monocular Depth Estimation Results for Aberration Images and GT Images}}

In this section, we present the depth estimation qualitative results, where the GT images and aberration images are predicted using the same metric Monocular Depth Estimation (MDE) model~\cite{piccinelli2024unidepth}.

\begin{figure*}[!t]

  \centering
  \includegraphics[width=1.0\linewidth]{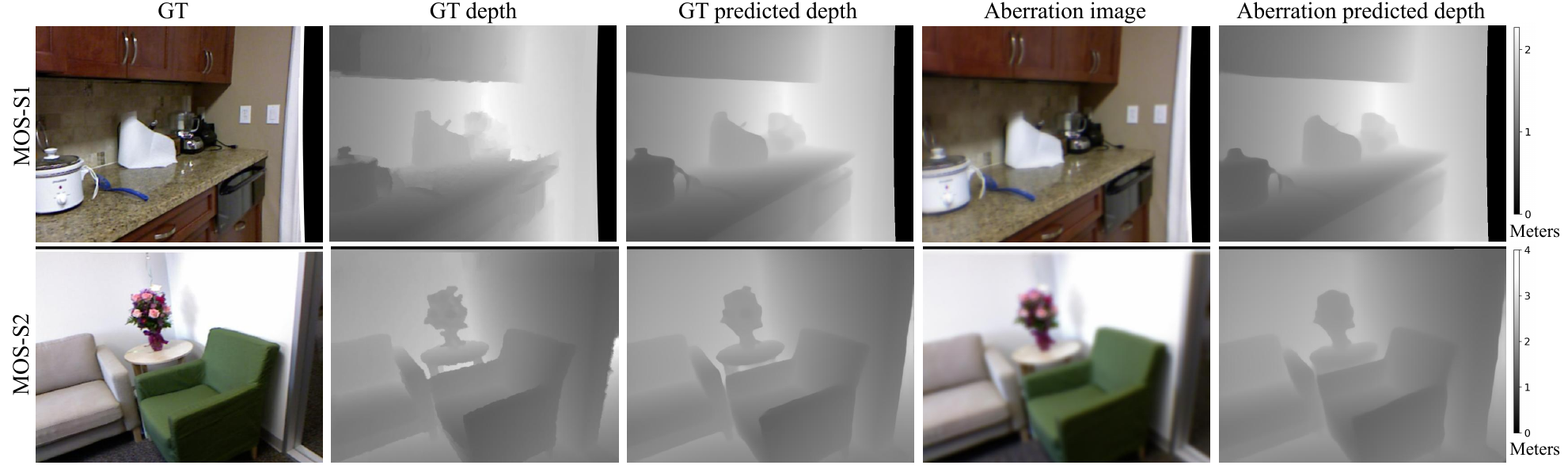}

  \caption{Qualitative comparison of MDE results of the aberration image and the GT image.}
  \label{fig:MDE_result}
\end{figure*}

\JQ{The qualitative results are shown in Fig.~\ref{fig:MDE_result}.} 
\JQ{The depth prediction results for both the aberration images and the GT images exhibit acceptable visual differences, with the depth predictions from the aberration images remaining reliable.}

\section{\JQ{Comparison of Aberration Correction Using Different Model Paradigms}}

\JQ{We compare the image restoration results of the SR-based method with those of defocus deblurring approaches~\cite{ruan2021aifnet,ruan2022learning}.}
\JQ{As shown in Table~\ref{tab:defocus_compare}, the SR-based method outperforms the defocus deblurring method in terms of metrics.}
\JQ{This advantage is rooted in findings from a previous study~\cite{jiang2023minimalist}, which demonstrated that the cylindrical network paradigm within the SR-based framework performs better on CAC tasks. Achieving high-quality image restoration results is essential for obtaining single-lens controllable DoF imaging.}
\JQ{Additionally, this work highlights the proposed training scheme and depth-aware mechanism, and our framework is not restricted to a specific network paradigm; other paradigms are also applicable.}

\begin{table*}[!t]
\centering
\caption{Quantitatively compare the aberration correction capabilities of the defocus deblurring method and our method.}
\label{tab:defocus_compare}
\renewcommand\arraystretch{1.2}
\huge
\resizebox{\textwidth}{!}{%
\begin{tabular}{|c|c|cccccc|cccccc|}
\hline
\multirow{3}{*}{Method} & \multirow{3}{*}{Pipeline} & \multicolumn{6}{c|}{MOS-S1}                                                                                                      & \multicolumn{6}{c|}{MOS-S2}                                                                                                      \\ \cline{3-14} 
                        &                           & \multicolumn{3}{c|}{NYUv2}                                                & \multicolumn{3}{c|}{MiddleBury2014}                  & \multicolumn{3}{c|}{NYUv2}                                                & \multicolumn{3}{c|}{MiddleBury2014}                  \\ \cline{3-14} 
                        &                           & PSNR$\uparrow$            & SSIM$\uparrow$           & \multicolumn{1}{c|}{LPIPS$\downarrow$}          & PSNR$\uparrow$            & SSIM$\uparrow$           & LPIPS$\downarrow$          & PSNR$\uparrow$            & SSIM$\uparrow$           & \multicolumn{1}{c|}{LPIPS$\downarrow$}          & PSNR$\uparrow$            & SSIM$\uparrow$           & LPIPS$\downarrow$          \\ \hline
AIFNet~\cite{ruan2021aifnet}                  & Depth-aware               & 37.7415          & 0.9715          & \multicolumn{1}{c|}{0.0373}          & 35.3338          & 0.9570          & 0.0763          & 34.1441          & 0.9388          & \multicolumn{1}{c|}{0.0745}          & 35.0936          & 0.9411          & 0.1055          \\ \hline
DRBNet~\cite{ruan2022learning}                  & Depth-aware               & 38.1294          & 0.9732          & \multicolumn{1}{c|}{0.0359}          & 35.4379          & 0.9587          & 0.0759          & 34.4226          & 0.9410          & \multicolumn{1}{c|}{0.0735}          & 34.9117          & 0.9427          & 0.1059          \\ \hline
RSTB~\cite{liang2021swinir}                    & Depth-aware               & \textbf{38.8280} & \textbf{0.9758} & \multicolumn{1}{c|}{\textbf{0.0342}} & \textbf{35.8148} & \textbf{0.9616} & \textbf{0.0736} & \textbf{34.8523} & \textbf{0.9450} & \multicolumn{1}{c|}{\textbf{0.0693}} & \textbf{35.3416} & \textbf{0.9446} & \textbf{0.1055} \\ \hline
\end{tabular}%
}
\end{table*}

\section{\JQ{Comparison of qualitative results of controllable DoF imaging}}

\begin{figure*}[!h]
  \centering
  \includegraphics[width=1.0\linewidth]{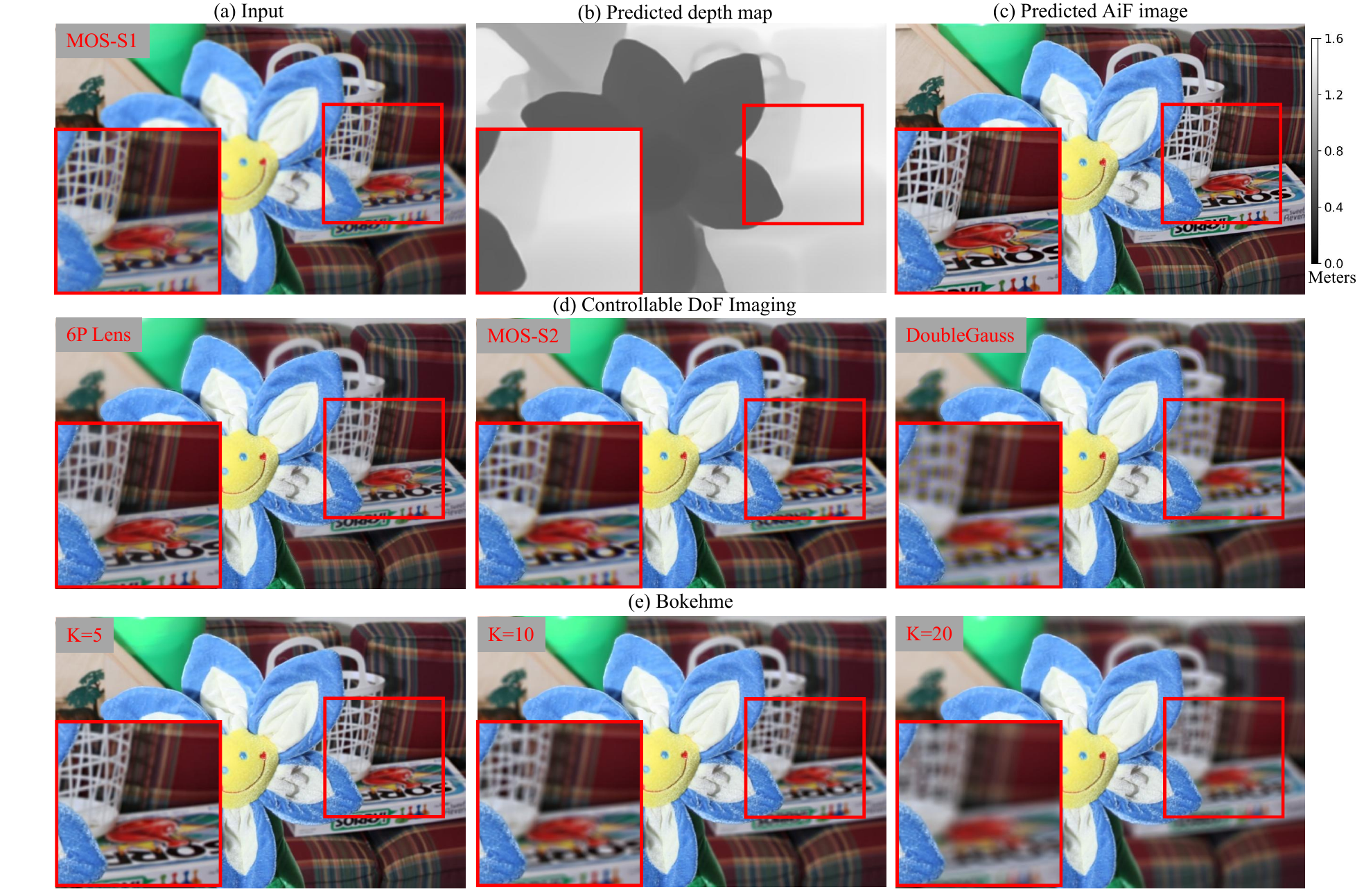}
  \caption{Comparison of qualitative results of controllable DoF imaging.}
  \label{fig:Dof_control_flower}
\end{figure*}

\JQ{To better demonstrate the rendering quality achieved by the proposed framework, we conducted a comparison with an existing rendering-based approach, BokehMe~\cite{peng2022bokehme}.
As shown in Fig.~\ref{fig:Dof_control_flower}, we first obtain the depth map of an aberration image through the metric MDE model~\cite{piccinelli2024unidepth} and then restore the image through the CAC module.}

\JQ{Fig.~\ref{fig:Dof_control_flower}(d) represents the controllable DoF imaging of different lenses generated by our proposed method.
As illustrated in Fig.~\ref{fig:Dof_control_flower}(e), BokehMe’s imaging results achieve different degrees of blur by adjusting the blur parameter $K$, and the blur radius can be calculated by
\begin{equation}
    r=K\left | d - d_{f}  \right | ,
\end{equation}
where $d$ denotes disparity (inverse depth) and $d_{f}$ represents the disparity of the focal plane.}

\JQ{
To facilitate fair comparison, we chose $K{=}\{5,10,20\}$ respectively to ensure that the blur level is similar to the three lenses represented by Omni-Lens-Field.
By closely examining the three images in the far-right column, compared to BokehME, which relies solely on intensity modulation based on blur radius, our method generates blur effects of different lenses by characterizing the PSF of the real lens. 
This method can accurately reflect the complex effects of optical aberrations (including dispersion).
}

\section{\JQ{Depth distribution of the used datasets}}

\begin{figure*}[!t]
  \centering
  \includegraphics[width=1.0\linewidth]{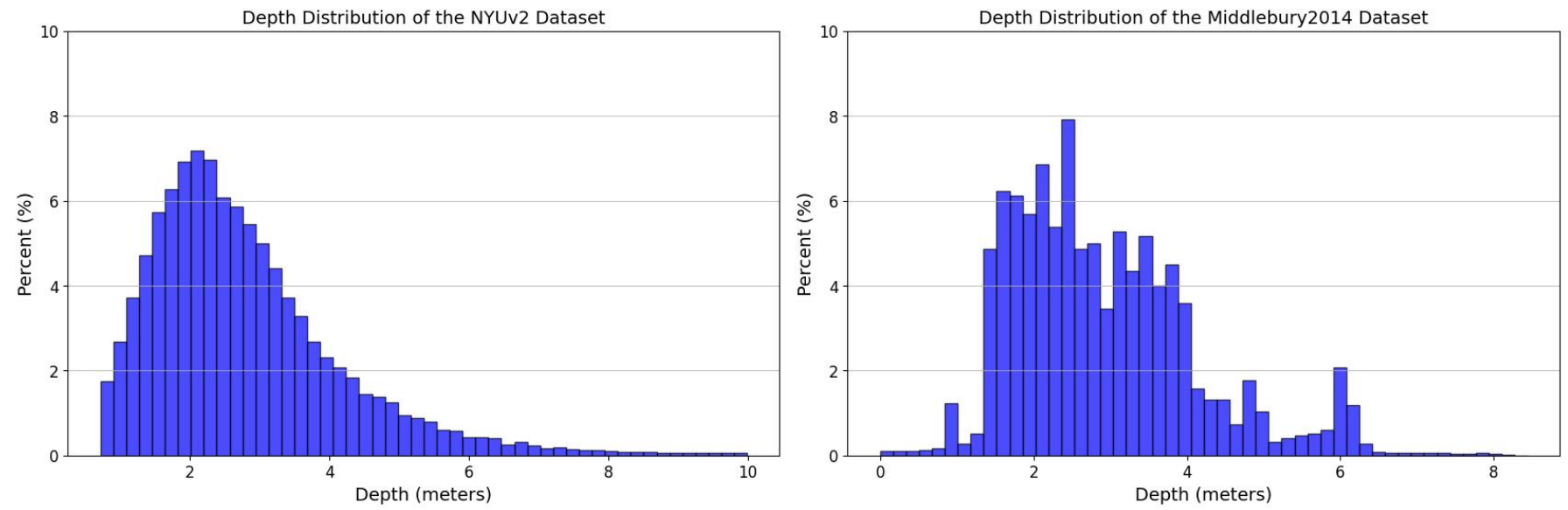}
  \caption{Depth distribution of the NYUv2 and Middlebury2014 datasets.}
  \label{fig:dataset range}
\end{figure*}

\JQ{
As shown in Fig.~\ref{fig:dataset range}, we show the depth distribution of the NYUv2~\cite{silberman2012indoor} and Middlebury2014~\cite{scharstein2014high} datasets.
Therefore, in this paper, the depth range of the test data is concentrated between $0.7m$ and $10m$ to obtain better comparison results.
}

\end{document}